\documentclass[11pt]{article}

\PassOptionsToPackage{table}{xcolor}
\usepackage[preprint]{acl}

\usepackage{times}
\usepackage{latexsym}

\usepackage[T1]{fontenc}

\usepackage[utf8]{inputenc}

\usepackage{microtype}

\usepackage{inconsolata}

\usepackage{graphicx}
\usepackage{amsmath}
\usepackage{amssymb}
\usepackage[ruled,vlined]{algorithm2e}
\usepackage{kotex}
\usepackage{booktabs}
\usepackage{tabularx}
\usepackage{multirow}
\usepackage{enumitem}
\usepackage{float}
\usepackage{stfloats}
\usepackage[most]{tcolorbox}
\usepackage{float}  
\usepackage{xcolor} 
\usepackage[table]{xcolor}
\usepackage{pifont}
\usepackage{adjustbox}
\usepackage{caption}
\usepackage{makecell}

\usepackage{cuted}
\usepackage{subcaption}
\newcommand{\bestcell}[1]{\cellcolor{red!24}\textbf{#1}}
\newcommand{\diag}[1]{\cellcolor{gray!15}#1}

\newcommand{\avgcell}[1]{\cellcolor{gray!8}{#1}}

\newcommand{\bestavg}[1]{\cellcolor{red!24}\textbf{#1}}

\usepackage[most]{tcolorbox}
\tcbuselibrary{skins,breakable}
\usepackage{xcolor}

\definecolor{promptblue}{HTML}{2563EB}
\definecolor{promptbg}{HTML}{F8FAFC}
\definecolor{promptline}{HTML}{BFDBFE}
\definecolor{prompttitle}{HTML}{EFF6FF}

\usepackage[most]{tcolorbox}
\tcbuselibrary{skins,breakable}

\definecolor{promptbg}{HTML}{F8FAFC}
\definecolor{promptframe}{HTML}{CBD5E1}
\definecolor{prompttitlebg}{HTML}{DBEAFE}
\definecolor{prompttitlefg}{HTML}{1E3A8A}

\usepackage[most]{tcolorbox}
\tcbuselibrary{skins,breakable}

\definecolor{promptbg}{HTML}{EFF6FF}       
\definecolor{promptframe}{HTML}{93C5FD}    
\definecolor{prompttitlebg}{HTML}{BFDBFE}  
\definecolor{prompttitlefg}{HTML}{1E3A8A}  

\newtcolorbox{promptbox}[1][]{
  enhanced,
  breakable,
  colback=promptbg,
  colframe=promptframe,
  boxrule=0.5pt,
  arc=4pt,
  outer arc=4pt,
  left=8pt,
  right=8pt,
  top=7pt,
  bottom=7pt,
  boxsep=0pt,
  before skip=6pt,
  after skip=6pt,
  title={#1},
  fonttitle=\bfseries\sffamily\small,
  coltitle=prompttitlefg,
  colbacktitle=prompttitlebg,
  boxed title style={
    colframe=prompttitlebg,
    colback=prompttitlebg,
    boxrule=0pt,
    arc=4pt,
    outer arc=4pt,
    left=7pt,
    right=7pt,
    top=3pt,
    bottom=3pt
  },
  attach boxed title to top left={
    xshift=5pt,
    yshift=-5pt
  },
  before upper={
    \small
    \setlength{\parindent}{0pt}
    \setlength{\parskip}{2pt}
  }
}

\newtcolorbox{gamelogbox}{
  breakable,
  colback=gray!8,
  colframe=gray!35,
  boxrule=0.4pt,
  arc=2pt,
  left=6pt,
  right=6pt,
  top=6pt,
  bottom=6pt,
  fontupper=\small
}

\def\tcb{\textcolor{blue}}

\title{MARBO: Relational Belief Grounding for LLM Agents\\ in Social Deduction Games}

\author{
  Yechan Hwang\textsuperscript{1,*} \quad
  Sangjun Bae\textsuperscript{1,*} \quad
  Jeongmo Kim\textsuperscript{1} \quad
  Sangwoo Bang\textsuperscript{1} \quad
  Seungyul Han\textsuperscript{1,\textdagger} \\
  {\normalsize\texttt{yechanhwang@unist.ac.kr}} \quad
  {\normalsize\texttt{bsjuntiger@unist.ac.kr}} \quad
  {\normalsize\texttt{jmkim@unist.ac.kr}} \\
  {\normalsize\texttt{sangwoobang@unist.ac.kr}} \quad
  {\normalsize\texttt{syhan@unist.ac.kr}}
}

\begin{document}
\raggedbottom
\maketitle
\footnotetext[1]{Graduate School of Artificial Intelligence, UNIST, Ulsan, South Korea. \textsuperscript{*}Equal contribution. \textsuperscript{\textdagger}Corresponding author: Seungyul Han.}
\setcounter{footnote}{1}

\begin{abstract}
Social deduction games (SDGs) require agents to reason under partial observability by maintaining relational beliefs about hidden roles and team alignments.
While recent LLM-agent approaches improve gameplay through prompting and preference optimization, they often optimize actions and in-game speech without explicitly grounding them in such beliefs. 
This frequently leads to strategically inconsistent behavior, especially for compact LLM agents. We introduce Multi-Agent Relational Belief Optimization (MARBO), a belief-grounded preference optimization framework that leverages relational beliefs to guide strategic decisions and in-game speech. 
MARBO provides preference feedback only when behaviors are supported by reliable relational beliefs and lead to strategically favorable social outcomes, encouraging more consistent learning under uncertainty. 
Experiments on representative SDGs show that MARBO enables compact LLM agents to consistently outperform existing baselines.
The Code is available on \href{https://pleasetakemeaway.github.io/MARBO/}{https://github.com/PleaseTakemeAway/MARBO}.
\end{abstract}

\section{Introduction}
\label{sec:introduction}

Recently, LLM agents have evolved beyond natural language generation~\citep{openai2023gpt4} into interactive agents capable of collaboration, negotiation~\citep{bakhtin2022cicero,park2023generative}, and decision making in dynamic environments~\citep{wang2023voyager}. 
Social deduction games (SDGs), such as \textit{Werewolf}~\citep{bailis2024werewolf}, \textit{Avalon}~\citep{light2023avalonbench}, and \textit{Among Us}~\citep{golechha2025among}, instantiate such a setting, and have emerged as an important testbed for evaluating long-horizon social reasoning of LLM agents under partial observability.

To succeed in SDGs, agents are required to infer relational belief such as hidden roles and opponents' strategies while strategically selecting actions and in-game speech to win the game~\citep{wu2024enhance}.
Prior work has improved LLM-agent behavior through prompting~\citep{wang2024recon}, reflection~\citep{shinn2023reflexion}, and preference optimization~\citep{xu2025learning,ye2025multi}. 
However, we observe that existing approaches often fail to reason coherently, resulting in goal-inconsistent actions such as revealing hidden roles or making self-defeating accusations.
These limitations become particularly severe for compact LLM agents, which struggle to maintain long-horizon conversational context.

\begin{figure*}[!t]
\centering
\includegraphics[width=\linewidth]{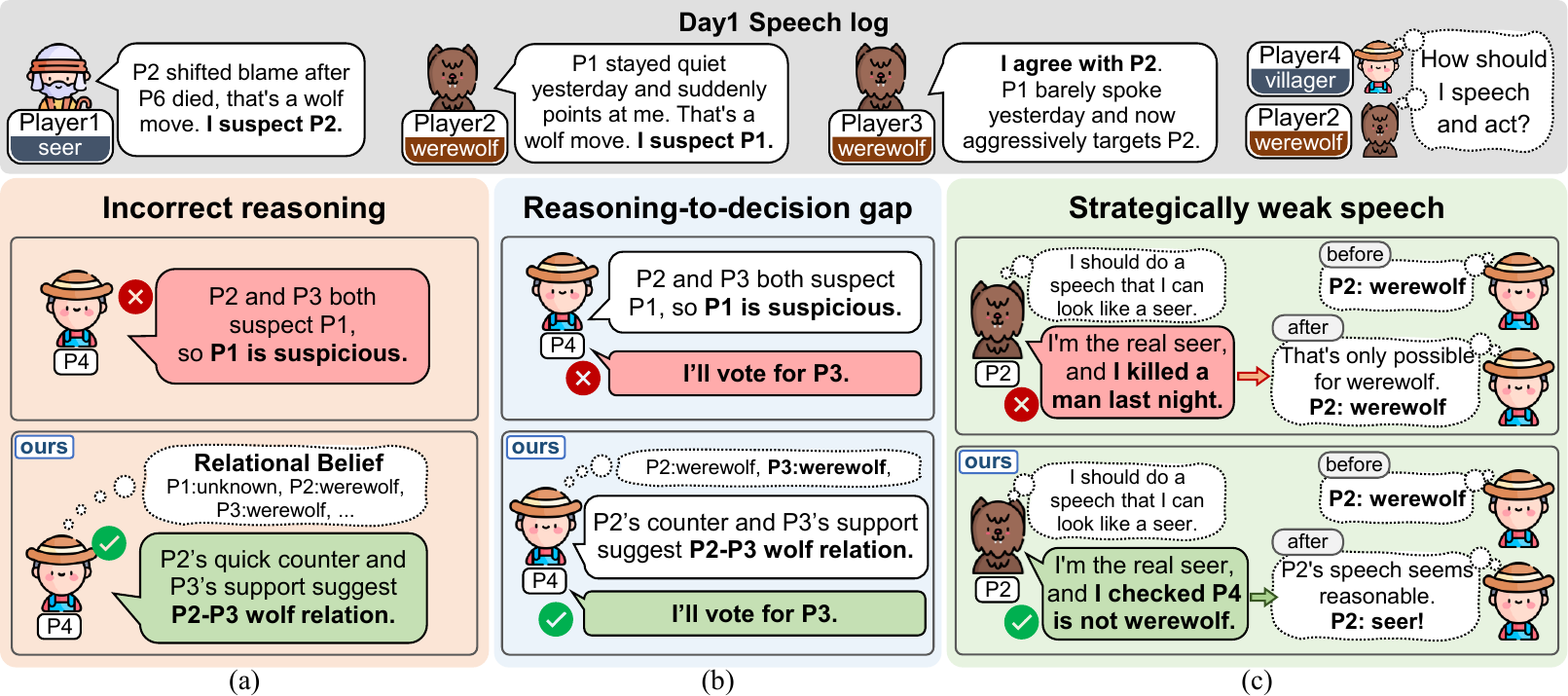}
\caption{Comparison of existing approaches and MARBO in \textit{Werewolf}:
(a) \textbf{Incorrect reasoning}: Existing approaches may follow surface-level accusations and wrongly suspect $P_1$, whereas MARBO infers a likely $P_2$--$P_3$ alliance.
(b) \textbf{Reasoning-to-decision gap}: Prior methods may fail to ground votes in inferred relations, while MARBO aligns decisions with relational beliefs.
(c) \textbf{Strategically weak speech}: Outcome-level optimization may produce weak speech, whereas MARBO generates belief-grounded speech that redirects suspicion.
}
\label{fig:fig1}
\end{figure*}

Figure~\ref{fig:fig1} illustrates several limitations of existing approaches in the \textit{Werewolf} game, where players discuss, vote, and eliminate opponents through night actions. In Figure~\ref{fig:fig1}(a), agents may rely on surface-level accusations rather than reasoning about latent player relations, often leading to incorrect reasoning and false suspicions about other players. Figure~\ref{fig:fig1}(b) further shows that voting decisions are often weakly grounded in the agent's own reasoning, causing inconsistent actions even when the underlying reasoning is partially correct. Finally, Figure~\ref{fig:fig1}(c) highlights that speech is typically optimized based on conversational quality or outcome-level feedback, often producing strategically weak speech that fails to effectively redirect suspicion.
We identify the root cause of these failures as the absence of \textit{relational belief grounding} during optimization, as existing approaches often reinforce behaviors even when reasoning about other players is inaccurate or strategically disadvantageous, resulting in unreliable social reasoning under partial observability. 

To address this challenge, we propose \textbf{Multi-Agent Relational Belief Optimization (MARBO)}, a preference optimization framework that incorporates relational beliefs into training. MARBO uses inferred relational beliefs to guide preference feedback and selectively reinforces decisions and in-game speech only when they are supported by reliable reasoning and lead to favorable social outcomes. Consequently, MARBO addresses the limitations in Figure~\ref{fig:fig1} by improving reasoning consistency, aligning decisions with inferred relations, and encouraging strategically effective speech. Experiments on representative games, including \textit{Werewolf} and \textit{Among Us}, demonstrate substantial performance gains, particularly for compact LLM agents. Our contributions are threefold:
\begin{itemize}[leftmargin=*, itemsep=0pt, topsep=0.2em]
\item \textbf{Relational belief formulation}. We formalize \textit{relational beliefs} as explicit representations of inter-player reasoning in partially observable SDGs, capturing hidden roles, alliances, suspicion, and player perceptions.
\item \textbf{Belief-grounded preference optimization}. We propose MARBO, a preference optimization framework that selectively reinforces strategic decisions and in-game speech only when they are supported by sufficiently reliable relational beliefs and lead to favorable social outcomes.
\item \textbf{Empirical effectiveness and analysis in SDGs}. We demonstrate that MARBO consistently improves compact LLM-agent performance on representative SDGs and provide extensive analysis showing how belief-grounded optimization improves strategic consistency and social reasoning.
\end{itemize}

\section{Related Work}
\label{sec:related_works}

\paragraph{Enhancing Reasoning in LLMs.}
LLM reasoning has advanced through prompting and structured search~\citep{yao2022react,yao2023tree,besta2024graph, bae2026llm}, experience-based reasoning~\citep{zhao2024expel,yang2024buffer}, and training-based supervision using learned verifiers~\citep{cobbe2021training}, process supervision~\citep{lightman2024let},
multi-agent self-play~\citep{liu2025spiral}, and multi-turn social interaction~\citep{jiang2026one}. 
Preference optimization aligns model behavior through RLHF~\citep{christiano2017deep,ouyang2022training,bai2022training}, likelihood calibration~\citep{zhao2023slic}, response ranking~\citep{yuan2023rrhf}, and preference sampling~\citep{liu2024statistical}, while DPO~\citep{rafailov2023dpo} and KTO~\citep{pmlr-v235-ethayarajh24a} enhance reasoning without explicit reward model.

\paragraph{Multi-agent Systems in Markov Games.}
Multi-agent systems have been extensively studied in Markov games. For example, multi-agent reinforcement learning has been widely applied to game environments involving multiple agents, such as \textit{StarCraft} and football, from various perspectives, including credit assignment~\citep{rashid2018qmix,yu2022mappo,kim2026gpae}, addressing partial observability~\citep{jo2024fox,jo2026retaining, bae2026llm}, and robustness~\citep{bukharin2023robust,pmlr-v267-lee25h,lee2026interaction}. Beyond these environments, coordination among agents has also been studied in cooperative games such as \textit{Overcooked} and \textit{Hanabi}~\citep{pmlr-v119-hu20a,pmlr-v139-lupu21a,kang2026shaping}. More recently, text-based SDGs, such as \textit{Werewolf} and \textit{Avalon}, have emerged as challenging multi-agent environments that require reasoning over hidden roles and intentions~\citep{light2023avalonbench,bailis2024werewolf,rahimirad2026bayesian}. LLM-based approaches to SDGs have evolved from prompt-based strategies with memory or reflection~\citep{xu2023exploring,wang2024recon} to learning-based methods for strategic decision-making and hidden-role inference~\citep{xu2024strategic,ye2025multi,zhang2025multimind}.

\paragraph{Relational Belief Modeling.}
Relational belief modeling concerns how an agent represents not only what others know, but how they stand toward one another. Prior work~\citep{gmytrasiewicz2005framework} has formalized social reasoning using explicit representations~\citep{stuhlmuller2014reasoning} and modeled Theory of Mind~\citep{baker2011bayesian} reasoning through nested probabilistic conditioning. Other approaches model relational states through explicit structured representations~\citep{hoorn2023silicon}, later extended to multidimensional decision making~\citep{ho2022quantum} and to observer-dependent inference conditioned on the observer's prior epistemic state~\citep{hoorn2026observer}.

Despite these advances, prior work typically uses inferred social information primarily as auxiliary reasoning signals rather than explicitly coupling it with behavior optimization. In contrast, MARBO incorporates relational beliefs into preference optimization, enabling strategic behavior and in-game speech to be optimized with relation-aware feedback.

\section{Preliminaries}
\label{sec:preliminaries}
\vspace{-0.2em}
\paragraph{Partially Observable Markov Games.}
We formalize SDGs as a partially observable Markov game (POMG), represented by
$
G=
\langle
N,
\mathcal{S},
\{\mathcal{A}^i\}_{i=1}^{N},
P,
R,
\mathcal{O},
H
\rangle,$
where \(N\) denotes the number of agents, \(\mathcal{S}\) the state space, \(\mathcal{A}^i\) the action space of agent \(i\), \(P\) the transition dynamics, \(R\) the reward function, \(\mathcal{O}\) the observation function, and \(H\) the episode horizon.

In language-based multi-agent environments, each agent action consists of two components: \textit{in-game speech} and \textit{task-specific actions}, denoted by \(a_{\mathrm{speech},t}^i\) and \(a_{\mathrm{task},t}^i\), respectively. Here, \(a_{\mathrm{speech},t}^i\) corresponds to natural-language interactions, while \(a_{\mathrm{task},t}^i\) represents one or more game-dependent actions, such as voting or role-specific actions (e.g., elimination, investigation, or protection). The state \(s_t\in\mathcal S\) contains both hidden social information and publicly observable game information.

At timestep \(t\), agent \(i\) receives a partial observation \(o_t^i=\mathcal O_i(s_t)\) and samples an action according to $
a_t^i
\sim
\pi_\theta^i
(
\cdot
\mid
\tau_t^i
),$ where \(\pi_\theta^i\) denotes the LLM policy parameterized by \(\theta\), and \(\tau_t^i=(I^i,h_t^i,o_t^i)\) represents the agent-specific context consisting of role-dependent instructions \(I^i\), interaction history (or its summary) \(h_t^i\), and the current observation \(o_t^i\). Let \(\mathbf a_t=(a_t^1,\ldots,a_t^N)\) denote the joint action of all agents. The environment transitions according to $
s_{t+1}
\sim
P(\cdot\mid s_t,\mathbf a_t),$ and the objective is to maximize the expected return $
\mathbb E
\left[
\sum_{t=0}^{H-1}
R(s_t,\mathbf a_t)
\right].$

\paragraph{Preference Optimization in SDGs.}
Beyond supervised fine-tuning (SFT), which imitates demonstrated behaviors, preference optimization has emerged as an effective paradigm for improving LLM reasoning and task performance using desirable and undesirable feedback. Among these approaches, Kahneman--Tversky Optimization (KTO) optimizes independently labeled desirable and undesirable responses.

Given a prompt-response pair \((x,y)\) with signed feedback \(d(x,y)\), where \(d(x,y)>0\) and \(d(x,y)<0\) indicate desirable and undesirable responses, respectively, KTO defines the utility as
\begin{equation}
v(x,y)=
\left\{
\begin{array}{@{}l@{\;}l@{}}
\lambda_D
\sigma
(
\beta
(
r_\theta(x,y)-z_0
)
),
&
d(x,y)>0,
\\
\lambda_U
\sigma
(
\beta
(
z_0-r_\theta(x,y)
)
),
&
d(x,y)<0,
\end{array}
\right.
\end{equation}
where
\[
r_\theta(x,y)
=
\log
\frac{
\pi_\theta(y\mid x)
}{
\pi_{\mathrm{ref}}(y\mid x)
}
\]
denotes the log-likelihood ratio between the current and reference policies, \(\sigma(\cdot)\) is the sigmoid function, \(z_0\) is the reference point, \(\beta\) controls sharpness, and \(\lambda_D,\lambda_U\) weight desirable and undesirable responses. The KTO objective is
\begin{equation}
\mathcal L_{\mathrm{KTO}}
=
\mathbb E_{(x,y)\sim\mathcal D}
[
\lambda_y-v(x,y)
],
\end{equation}
where \(\lambda_y=\lambda_D\) if \(d(x,y)>0\), and \(\lambda_y=\lambda_U\) otherwise.

Preference optimization is particularly suitable for SDGs, where different interaction phases can be optimized independently. Following this idea, MaKTO applies KTO to phase-specific behaviors for training LLM agents in SDGs. We adopt a similar optimization framework but additionally incorporate \textit{relational beliefs} to construct belief-grounded preferences for social reasoning across both speech and task actions.

\section{Methodology}
\label{sec:methodology}

\begin{figure*}[t]
\vspace{-1.5em}
\centering
\includegraphics[width=\linewidth]{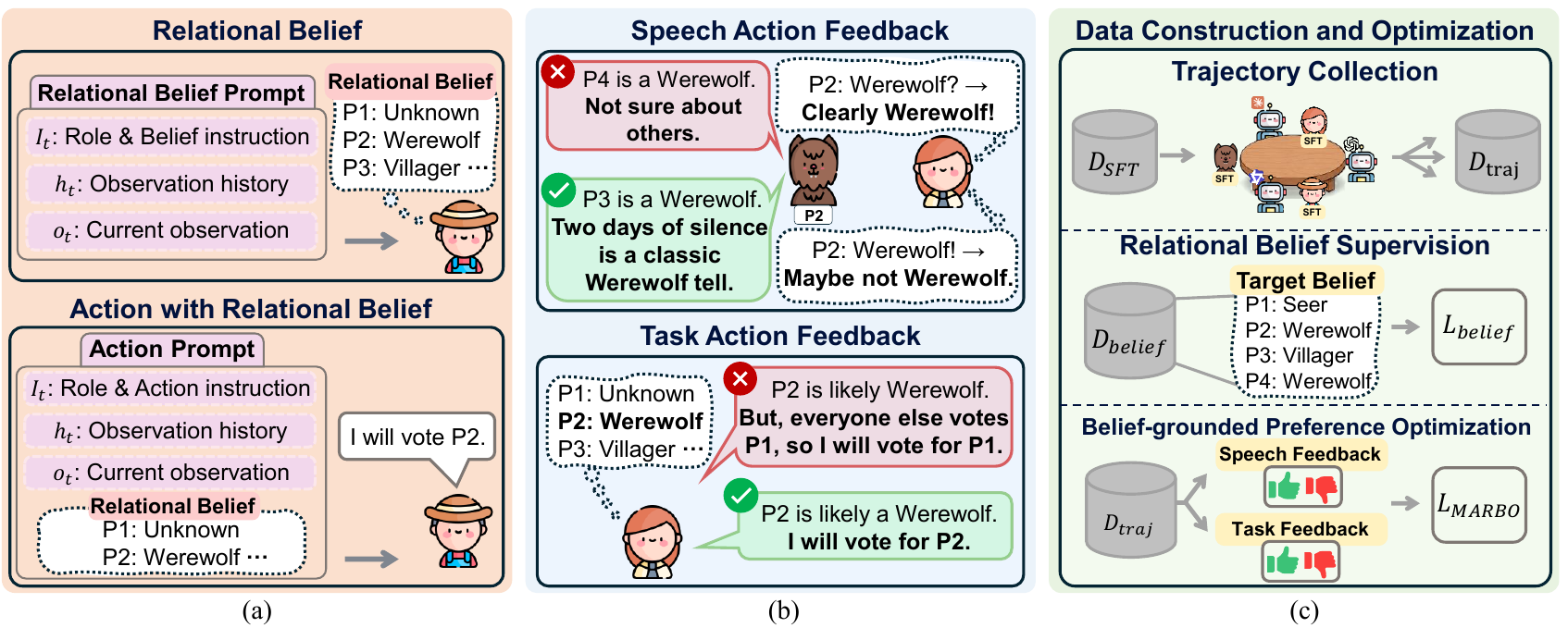}
\caption{Overview of MARBO. (a) Each agent infers relational beliefs over other players' hidden social labels and conditions action generation on them.
(b) MARBO constructs belief-grounded feedback for speech and task actions: speech feedback combines $\eta_{\mathrm{speech}}^{\mathrm{LLM}}$ with favorable belief shifts, while task feedback combines $\eta_{\mathrm{task}}$ with target-belief correctness. (c) presents the overall MARBO training framework.}
\label{fig:method}
\end{figure*}

\subsection{Relational Belief Formulation}
\label{subsec:relational_belief}

To address failures caused by partial observability, MARBO explicitly models how agents infer other players' hidden social identities through \textit{relational beliefs}. In SDGs, strategic behavior depends on reasoning about latent social information, such as hidden roles or team affiliations, which cannot be directly observed.

For each agent \(i\), let \(\mathcal{C}\) denote the possible social-label set, \(c^j \in \mathcal{C}\) the ground-truth label of agent \(j\), and \(\mathbf{c}=(c^1,\ldots,c^N)\) the joint ground-truth labels. 
Depending on the environment, social labels correspond to hidden roles or team affiliations. 
We define the relational belief of agent \(i\) about agent \(j\) at timestep \(t\) as
\begin{equation}
\hat b_t^{ij}
\sim
\pi_\theta^i
(
\cdot
\mid
\tau_t^i
),
\quad
\hat b_t^{ij}
\in
\boldsymbol{\mathcal{C}}
\cup
\{
\texttt{unknown}
\},
\quad
 j\neq i,
\end{equation}
where the additional \texttt{unknown} label captures uncertainty when the agent cannot reliably infer another player's social identity.

\begin{table}[H]
\centering
\small
\caption{Examples of social label sets across SDGs.}
\label{tab:belief_examples}
\vspace{-0.5em}
\begin{tabular}{ll}
\toprule
Game & Social label set $\mathbf{\mathcal{C}}$\\
\midrule
\textit{Werewolf}
&
$
\{
\texttt{Werewolf},
\texttt{Villager},
\texttt{Seer},
\ldots
\}$
\\
\textit{Among Us}
&
$
\{
\texttt{Crewmate},
\texttt{Impostor}
\}$
\\
\bottomrule
\end{tabular}
\vspace{-0.8em}
\end{table}

Examples of social labels are summarized in Table~\ref{tab:belief_examples}. In \textit{Werewolf}, social labels correspond to hidden roles, whereas in \textit{Among Us}, they correspond to team affiliations. Agent \(i\) conditions actions on inferred relational beliefs:
\begin{equation}
a_t^i
\sim
\pi_\theta^i
(
\cdot
\mid
\tau_t^i,
\hat b_t^{i1},
\ldots,
\hat b_t^{iN}
).
\end{equation}

\subsection{Belief-grounded Preference Design}
\label{subsec:belief_preference}

Existing approaches often reinforce desirable behaviors without verifying whether the underlying reasoning about other players is correct. As a result, agents may receive positive feedback even when their inferred understanding of others is inaccurate, leading to unreliable strategic behavior under partial observability. To address this issue, MARBO introduces a belief-grounded optimization framework consisting of two components: (1) relational belief supervision and (2) belief-grounded preference design.

\paragraph{(1) Relational belief supervision.}
Since accurate reasoning about other players is essential in SDGs, MARBO first trains agents to infer other players' social identities through supervised learning. Specifically, we optimize the following belief prediction objective:
\begin{equation}
\mathcal L_{\mathrm{belief}}^i
=
\mathbb E_{
(\tau_t^i,\mathbf{c})
\sim
\mathcal D_{\mathrm{belief}}
}
\left[
-\log
\pi_\theta^i
(
\mathbf{c}
\mid
\tau_t^i
)
\right],
\label{eq:belief_loss}
\end{equation}
where \(\mathcal D_{\mathrm{belief}}\) denotes the trajectory dataset for relational belief supervision. This objective improves relational belief inference about other players. Details of belief-label construction are provided in Appendix~\ref{app:label_construction}.

\paragraph{(2) Belief-grounded preference construction.}
Simply rewarding favorable outcomes can introduce misleading supervision when the underlying relational reasoning is incorrect. To address this issue, MARBO constructs preferences using both inferred beliefs and action types, distinguishing between \textit{speech actions} and \textit{task actions}. This allows preference signals to reflect not only whether an action appears successful, but also whether it is grounded in reliable relational reasoning.

\paragraph{Speech-action preference.}
Rather than evaluating speech solely based on conversational plausibility or outcome-level feedback, MARBO evaluates speech through its effect on other players' relational beliefs. This discourages strategically harmful speech, such as exposing hidden roles or increasing suspicion toward oneself.

Let \(\hat b_t^{ji}\) denote agent \(j\)'s relational belief about agent \(i\) before agent \(i\) performs speech action \(a_{\mathrm{speech},t}^i\). After the speech, we separately infer the updated belief of agent \(j\) using the updated interaction context:
\begin{equation}
\hat b_{t,+}^{ji}
\sim
\pi_\theta^j
(
\cdot
\mid
\tau_t^j,
a_{\mathrm{speech},t}^i
),
\end{equation}
where \(\hat b_{t,+}^{ji}\) denotes the updated belief immediately after the speech.

Following prior work, we first assess speech validity using an LLM-based judge\footnote{The LLM judge is implemented using the API-based \texttt{claude-sonnet-4-5-20250929}.}
$
\eta_{\mathrm{speech}}^{\mathrm{LLM}}
(
\tau_t^i,
a_{\mathrm{speech},t}^i
),
$ which evaluates whether speech is contextually valid and non-hallucinatory, independent of relational beliefs. We then define a belief-shift function
\[
\delta
(
a_{\mathrm{speech},t}^i,
\hat b_t^{ji},
\hat b_{t,+}^{ji}
)
\in
\{-1,0,+1\},
\]
where \(\delta=0\) if no belief change occurs, \(\delta=+1\) if the speech is valid according to \(\eta_{\mathrm{speech}}^{\mathrm{LLM}}\) and shifts agent \(j\)'s belief in a strategically favorable direction for agent \(i\), and \(\delta=-1\) otherwise. Strategic favorability is defined according to the game objective. For example, in voting-based games such as \textit{Werewolf} and \textit{Among Us}, being identified as a \texttt{Werewolf} or \texttt{Impostor} by a majority of players is strategically harmful. Hence, belief shifts that reduce such suspicion are treated as favorable. Details of the LLM-judge criteria and game-specific definitions of strategic favorability are provided in Appendix~\ref{app:reward_function}.

We then aggregate belief shifts across players and define the speech feedback score as
\begin{equation}
d_{\mathrm{speech},t}^{i}
=
\sum_{j\neq i}
\delta
(
a_{\mathrm{speech},t}^i,
\hat b_t^{ji},
\hat b_{t,+}^{ji}
).
\label{eq:speech_feedback1}
\end{equation}

According to Eq.~\eqref{eq:speech_feedback1}, speech is considered desirable when the aggregated belief shift is positive and undesirable otherwise. For the games considered in this work, adversarial agents (e.g., \texttt{Werewolf} or \texttt{Impostor}) already know their teammates. Therefore, belief shifts are computed only with respect to non-adversarial players (e.g., \texttt{Villager}, \texttt{Crewmate}).

\paragraph{Task-action preference.}
Unlike speech actions, task actions directly affect the game outcome and therefore require accurate target selection. Existing approaches often reward favorable outcomes regardless of whether the underlying reasoning about other players is correct, which can reinforce incorrect reasoning and induce strategically inconsistent behaviors. To address this issue, MARBO assigns desirable preferences only when a task action both aligns with the game objective of the corresponding social label and is grounded in correct relational beliefs.

To determine whether a task action aligns with the game objective, we first define a game-dependent task utility function
\[
\eta_{\mathrm{task}}
(
a_{\mathrm{task},t}^i,
\mathbf{c}
)
\in
\{-1,+1\},
\]
where $+1$ and $-1$ indicate whether the action is strategically favorable under the game rule, respectively. Here, favorability is determined using the ground-truth social labels $\mathbf{c}$.
For example, in \textit{Werewolf}, a \texttt{Villager} voting for a \texttt{Werewolf} is considered favorable, since eliminating \texttt{Werewolf} players aligns with the \texttt{Villager} players' objective. 
Details of game-specific task utility are provided in Appendix~\ref{app:task_utility}.

To avoid indiscriminate preference assignment and encourage agents to correctly identify the intended target of a task action, let \(j=\mathrm{tar}(a_{\mathrm{task},t}^i)\) denote the target player of task action \(a_{\mathrm{task},t}^i\).

We define task-action feedback as
\begin{equation}
d_{\mathrm{task},t}^{i}
=
\begin{cases}
\eta_{\mathrm{task}}
(
a_{\mathrm{task},t}^{i},
\mathbf{c}
),
&
\hat b_t^{ij}
=
c^j,
\\
-1,
&
\text{otherwise}.
\end{cases}
\label{eq:task_feedback}
\end{equation}
Most SDGs considered in this work involve task actions with explicit target players, making the feedback well-defined. For actions without explicit targets, feedback is determined solely by \(\eta_{\mathrm{task}}\). This design ensures that agents receive positive feedback only when they correctly infer the target player's social identity and execute a favorable task action. Consequently, MARBO discourages belief-inconsistent decisions and improves decision reliability under partial observability.

\subsection{Data Construction and Optimization}
\label{subsec:optimization}

Based on the proposed belief-grounded preferences, MARBO trains the LLM policy \(\pi_\theta\) through KTO-based optimization. 
We first perform supervised fine-tuning (SFT) using the environment-provided training dataset \(D_{\mathrm{SFT}}\). 
Following prior work, we then collect a gameplay trajectory dataset \(D_{\mathrm{traj}}\) using a diverse LLM model pool consisting of the SFT model and additional LLM agents assigned to different game roles. 
During trajectory collection, assigned agents interact in the game under their respective policy configurations. 
We store the resulting trajectories together with observable interaction histories and ground-truth social labels for relational belief supervision and preference construction.

Using the collected dataset, MARBO jointly optimizes KTO-based preference learning and relational belief supervision:
\begin{equation}
\mathcal L_{\mathrm{MARBO}}(\theta)
=
\mathcal L_{\mathrm{KTO}}(\theta)
+
\lambda_{\mathrm{belief}}
\mathcal L_{\mathrm{belief}}(\theta),
\label{eq:total_loss}
\end{equation}
where \(\lambda_{\mathrm{belief}}\) controls the strength of relational belief supervision. 
We use the full trajectory dataset, \(D_{\mathrm{belief}}=D_{\mathrm{traj}}\), for relational belief supervision, while preference optimization is performed only on trajectories generated by the SFT-initialized base model. 
Figure~\ref{fig:method} illustrates (a) relational belief inference and belief-conditioned action generation, (b) belief-grounded preference construction for speech and task actions, and (c) the overall MARBO framework. 
Additional implementation details are provided in Appendix~\ref{app:implementation_details}.

\section{Experiments}
\label{sec:experiments}

\subsection{Experimental Setup}

We evaluate MARBO on \textit{Werewolf} and \textit{Among Us} under side-balanced fixed-opponent and head-to-head settings. For \textit{Werewolf}, we run 100 games in a 9-player Seer-Guard-Witch setting with 3 Werewolves and 6 Villager-side players. For \textit{Among Us}, we run 200 games in a 7-player setting with 5 Crewmates and 2 Impostors. 
We use CoT-style outputs that are decomposed into the intended role and the speech to measure the resulting belief shift. 
Further details are provided in Appendix~\ref{app:prompt}.  

\paragraph{Models and baselines.}
For comparison, we use base models Qwen2.5 (7B/14B) for \textit{Werewolf} and Gemma4 (E2B/E4B) for \textit{Among Us}, both without game-specific SFT, to examine the dependence on SFT initialization. 
We compare MARBO against \textbf{Base}, the corresponding backbone or SFT-initialized model \textbf{ReCon}, a prompting-based agent with structured reasoning and relational memory; and \textbf{MaKTO}, a multi-agent KTO baseline trained on interaction trajectories. 
For fixed-opponent evaluation, we use API-based LLM opponents, including GPT-4o-mini, GPT-4o, and Claude-4.5-Haiku.\footnote{We use API model versions \texttt{gpt-4o-mini-2024-07-18}, \texttt{gpt-4o-2024-08-06}, and \texttt{claude-haiku-4-5-20251001}.} Additional implementation details are provided in Appendix~\ref{app:implementation_details}.

\paragraph{Training datasets.}
For \textit{Werewolf} and \textit{Among Us}, we construct offline datasets containing 3.5k and 5k-samples $D_{traj}$ from 300 randomly played games. 
Each sample is a decision-level input-output pair with signed feedback derived as described in Section~\ref{subsec:belief_preference}, covering speech actions, environment-specific task decisions, and belief-grounding samples. Further details, including feedback distributions and agent pools, are provided in Appendix~\ref{app:dataset_details}.

To evaluate MARBO, we focus on four questions:
\begin{itemize}[itemsep=0.1em, topsep=0.2em, parsep=0pt, partopsep=0pt]
    \item \textbf{Q1.} Does MARBO improve overall performance across SDGs and opponent models?
    \item \textbf{Q2.} Does MARBO learn accurate relational beliefs under partial observability?
    \item \textbf{Q3.} Does belief-shaping feedback improve strategic communication?
    \item \textbf{Q4.} Does belief-grounded feedback improve target-directed task actions?
\end{itemize}
Together, these questions assess both win-rate gains and how relational beliefs guide decisions and communication.

\begin{table*}[t]
\vspace{-1em}
\centering
\captionsetup{skip=2pt}
\small
\setlength{\tabcolsep}{6pt}
\renewcommand{\arraystretch}{1.12}
\caption{
Main results on Werewolf and Among Us, reporting mean win rate (\%) $\pm$ standard deviation across seeds.
}
\label{tab:main_results_combined}

\resizebox{\textwidth}{!}{
\begin{tabular}{llcccccc}
\toprule
\multicolumn{8}{l}{\cellcolor{blue!8}\textbf{Werewolf} \quad \textit{9-player Seer-Guard-Witch setting}} \\
\midrule
\textbf{Size}
& \textbf{Method}
& \textbf{GPT-4o-mini}
& \textbf{GPT-4o}
& \textbf{Claude-4.5-Haiku}
& \textbf{Gemma4-31B}
& \textbf{Qwen3.6-27B}
& \textbf{Avg. Win Rate}
\\
\midrule
\multirow{4}{*}{7B}
& Base
& $46.6_{\pm 3.7}$ & $40.4_{\pm 4.2}$ & $25.2_{\pm 4.6}$ & $23.8_{\pm 3.9}$ & $29.8_{\pm 4.4}$ & \avgcell{$33.2_{\pm 2.1}$} \\
& ReCon
& $43.8_{\pm 4.1}$ & $21.8_{\pm 4.8}$ & $17.8_{\pm 3.6}$ & $29.0_{\pm 4.5}$ & $24.2_{\pm 4.0}$ & \avgcell{$27.3_{\pm 0.9}$} \\
& MaKTO
& $52.2_{\pm 4.3}$ & $46.8_{\pm 3.8}$ & $45.0_{\pm 4.7}$ & $33.8_{\pm 4.2}$ & $30.2_{\pm 3.9}$ & \avgcell{$41.6_{\pm 1.6}$} \\
& Ours
& \bestcell{$\mathbf{70.2_{\pm 2.6}}$} & \bestcell{$\mathbf{67.8_{\pm 2.5}}$} & \bestcell{$\mathbf{64.0_{\pm 2.1}}$} & \bestcell{$\mathbf{61.0_{\pm 2.7}}$} & \bestcell{$\mathbf{65.2_{\pm 2.4}}$} & \bestavg{$\mathbf{65.6_{\pm 1.7}}$} \\
\cmidrule(lr){1-8}
\multirow{4}{*}{14B}
& Base
& $73.8_{\pm 4.5}$ & $56.2_{\pm 3.8}$ & $52.0_{\pm 4.7}$ & $50.0_{\pm 4.1}$ & $58.0_{\pm 3.9}$ & \avgcell{$58.0_{\pm 0.9}$}\\
& ReCon
& $69.8_{\pm 4.3}$ & $62.0_{\pm 4.6}$ & $56.0_{\pm 3.7}$ & $61.0_{\pm 4.8}$ & $62.0_{\pm 4.0}$ & \avgcell{$62.2_{\pm 1.6}$} \\
& MaKTO
& $75.0_{\pm 3.9}$ & $63.0_{\pm 4.4}$ & $58.0_{\pm 4.6}$ & $53.0_{\pm 3.8}$ & $68.0_{\pm 4.7}$ & \avgcell{$63.4_{\pm 1.8}$} \\
& Ours
& \bestcell{$\mathbf{80.2_{\pm 2.4}}$} & \bestcell{$\mathbf{74.8_{\pm 2.8}}$} & \bestcell{$\mathbf{66.2_{\pm 2.2}}$} & \bestcell{$\mathbf{68.0_{\pm 2.6}}$} & \bestcell{$\mathbf{72.0_{\pm 2.0}}$} & \bestavg{$\mathbf{72.2_{\pm 0.8}}$}  \\
\midrule
\addlinespace[0.35em]

\multicolumn{8}{l}{\cellcolor{orange!10}\textbf{Among Us}} \\
\midrule
\textbf{Size}
& \textbf{Method}
& \textbf{Qwen2.5-14B}
& \textbf{Qwen2.5-32B}
& \textbf{Phi-4}
& \textbf{Gemma4-31B}
& \textbf{Qwen3.6-27B}
& \textbf{Avg. Win Rate}
\\
\midrule
\multirow{4}{*}{E2B}
& Base
& $36.0_{\pm 5.7}$ & $24.0_{\pm 4.4}$ & $38.0_{\pm 1.4}$ & $13.0_{\pm 2.0}$ & $14.0_{\pm 5.5}$ & \avgcell{$25.0_{\pm 3.8}$} \\
& ReCon
& $48.0_{\pm 4.7}$ & $32.0_{\pm 2.6}$ & $47.0_{\pm 3.7}$ & $19.0_{\pm 3.8}$ & $30.0_{\pm 3.0}$ & \avgcell{$35.2_{\pm 3.6}$} \\
& MaKTO
& $47.0_{\pm 4.8}$ & $35.0_{\pm 3.0}$ & $46.0_{\pm 2.0}$ & $20.0_{\pm 3.9}$ & $34.0_{\pm 3.9}$ & \avgcell{$36.4_{\pm 3.5}$} \\
& Ours
& \bestcell{$\mathbf{70.0_{\pm 3.3}}$} & \bestcell{$\mathbf{49.0_{\pm 2.8}}$} & \bestcell{$\mathbf{55.0_{\pm 2.6}}$} & \bestcell{$\mathbf{31.0_{\pm 2.0}}$} & \bestcell{$\mathbf{42.0_{\pm 3.5}}$} & \bestavg{$\mathbf{49.4_{\pm 2.8}}$} \\
\cmidrule(lr){1-8}
\multirow{4}{*}{E4B}
& Base
& $49.0_{\pm 5.7}$ & $41.0_{\pm 4.7}$ & $60.0_{\pm 2.7}$ & $24.0_{\pm 4.7}$ & $29.0_{\pm 3.5}$ & \avgcell{$41.0_{\pm 4.3}$} \\
& ReCon
& $42.0_{\pm 5.0}$ & $28.0_{\pm 3.1}$ & $44.0_{\pm 5.2}$ & $20.0_{\pm 4.8}$ & $26.0_{\pm 2.8}$ & \avgcell{$32.0_{\pm 4.2}$} \\
& MaKTO
& $47.0_{\pm 4.2}$ & $46.0_{\pm 3.5}$ & $50.0_{\pm 1.9}$ & $21.0_{\pm 3.0}$ & $29.0_{\pm 2.5}$ & \avgcell{$38.6_{\pm 3.0}$} \\
& Ours
& \bestcell{$\mathbf{73.0_{\pm 3.1}}$} & \bestcell{$\mathbf{71.0_{\pm 3.3}}$} & \bestcell{$\mathbf{69.0_{\pm 2.3}}$} & \bestcell{$\mathbf{30.0_{\pm 3.3}}$} & \bestcell{$\mathbf{51.0_{\pm 2.4}}$} & \bestavg{$\mathbf{59.0_{\pm 2.9}}$} \\
\bottomrule
\end{tabular}
}
\end{table*}

\subsection{Overall Strategic Performance}
\label{subsec:performance}

To address \textbf{Q1}, Table~\ref{tab:main_results_combined} reports win rates against five fixed opponent models on \textit{Werewolf} and \textit{Among Us}. 
MARBO consistently achieves the best average performance across both games and backbone sizes, improving over MaKTO by $+24$ and $+9$ percentage points on Qwen2.5-7B/14B in \textit{Werewolf} and by $+13$ and $+20$ percentage points on Gemma-E2B/E4B in \textit{Among Us}. 
A key observation is that MARBO-7B outperforms MaKTO-14B, indicating that the improvement is driven by belief-grounded supervision rather than simply by increasing model capacity. 
Moreover, the gains are consistent across all five opponents in both games, suggesting that MARBO learns generally useful strategic behavior instead of exploiting opponent-specific weaknesses.

\begin{table}[!t]
\centering

\captionsetup{skip=2pt}
\small
\caption{Head-to-head competition win rate(\%).}
\label{tab:head_to_head_combined}

\setlength{\tabcolsep}{2.5pt}
\renewcommand{\arraystretch}{1.13}

\begin{adjustbox}{max width=\columnwidth}
\begin{tabular}{lcccc}
\toprule
\textbf{Model} & \textbf{Base} & \textbf{ReCon} & \textbf{MaKTO} & \textbf{Ours} \\
\midrule

\rowcolor{blue!8}
\multicolumn{5}{l}{\textbf{Werewolf}} \\
Base & \diag{-} & $57.3_{\pm 5.1}$ & $39.3_{\pm 4.5}$ & $30.3_{\pm 3.1}$ \\
ReCon & $42.7_{\pm 5.1}$ & \diag{-} & $48.3_{\pm 5.1}$ & $35.7_{\pm 2.1}$ \\
MaKTO & $60.7_{\pm 4.5}$ & $51.7_{\pm 5.1}$ & \diag{-} & $41.7_{\pm 2.5}$ \\
\textbf{Ours} & \bestcell{$\mathbf{69.7_{\pm 3.1}}$} & \bestcell{$\mathbf{64.3_{\pm 2.1}}$} & \bestcell{$\mathbf{58.3_{\pm 2.5}}$} & \diag{-} \\

\addlinespace[0.45em]

\rowcolor{orange!10}
\multicolumn{5}{l}{\textbf{Among Us}} \\
Base & \diag{-} & $35.0_{\pm 2.3}$ & $33.0_{\pm 1.9}$ & $21.0_{\pm 3.0}$ \\
ReCon & $65.0_{\pm 2.3}$ & \diag{-} & $49.0_{\pm 2.8}$ & $27.0_{\pm 3.1}$ \\
MaKTO & $67.0_{\pm 1.9}$ & $51.0_{\pm 2.8}$ & \diag{-} & $31.0_{\pm 2.7}$ \\
\textbf{Ours} & \bestcell{$\mathbf{79.0_{\pm 3.0}}$} & \bestcell{$\mathbf{73.0_{\pm 3.1}}$} & \bestcell{$\mathbf{69.0_{\pm 2.7}}$} & \diag{-} \\
\bottomrule
\end{tabular}
\end{adjustbox}

\end{table}

Table~\ref{tab:head_to_head_combined} further evaluates this effect in a pairwise head-to-head setting against trained Base, ReCon, and MaKTO in 7B/E2B models. 
MARBO remains the strongest agent, achieving win rates of $70\%, 64\%, 58\%$ in \textit{Werewolf} and $79\%, 73\%, 69\%$ in \textit{Among Us} against Base/ReCon/MaKTO, respectively. 
These results support our intuition that explicitly grounding optimization in relational beliefs improves strategic consistency under partial observability, yielding gains that transfer across games, backbone sizes, and opponent distributions. 
Role-specific win rates are provided in Appendix~\ref{app:additional_experiments}.

\begin{figure}[!t]
    \centering
    \vspace{0.5em}
    \includegraphics[width=\columnwidth]{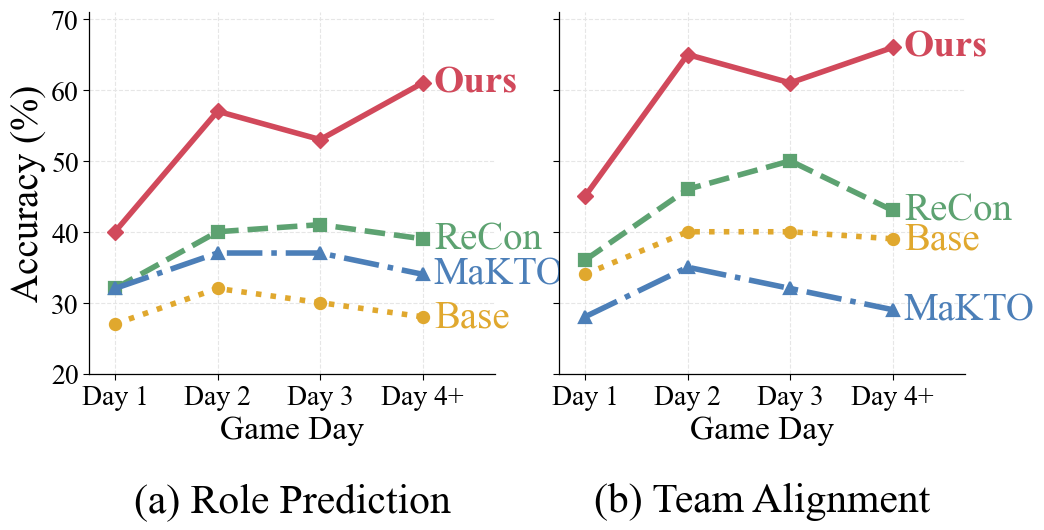}
    \captionsetup{skip=0.3em}
    \caption{Day-wise relational belief inference,(a) role prediction and (b) team alignment accuracy.}
    \label{fig:long_horizon_reasoning}
    \vspace{-0.3em}
\end{figure}

\subsection{Analysis of Belief-Grounded Behavior}
\label{subsec:belief_grounded_behavior_analysis}

For \textbf{Q2-Q4}, we evaluate Qwen2.5-7B over 5 seeds with 100 \textit{Werewolf} games per seed against GPT-4o-mini. 
We assess the belief-grounded behavior pipeline: relational belief construction, target-directed task actions, and belief-shaping communication. 
Additional \textit{Among Us} analyses and case studies appear in Appendix~\ref{app:case_study} and Appendix~\ref{app:additional_experiments}.

\paragraph{Relational belief construction.}
To answer \textbf{Q2}, we evaluate whether MARBO constructs accurate relational beliefs from partial observations using Alignment Accuracy for ally-enemy identification, Role Prediction Accuracy for overall role inference, and Wolf F1 for reliable werewolf detection that accounts for both false accusations and missed werewolves. 
Further details are provided in Appendix~\ref{subsecapp:relational_belief_experiments_details}.

\begin{table*}[t]
\centering
\captionsetup{skip=2pt}
\normalsize
\setlength{\tabcolsep}{4pt}
\renewcommand{\arraystretch}{1.15}
\caption{Belief-level evaluation. (a) relational belief construction accuracy, (b) role-wise role prediction accuracy,
and (c) belief-shaping success rate, measured as the accuracy (\%) with which opponents infer the speaker as its intended role.}
\label{tab:belief_all}
\newcolumntype{R}{>{\centering\arraybackslash}p{3.6em}}
\resizebox{\textwidth}{!}{%
\begin{tabular}{@{}l ccc @{\hspace{10pt}} RRRR @{\hspace{10pt}} ccc@{}}
\toprule
& \multicolumn{3}{c@{\hspace{10pt}}}{\textbf{(a) Relational belief}}
& \multicolumn{4}{c@{\hspace{10pt}}}{\textbf{(b) Role-wise role prediction accuracy}}
& \multicolumn{3}{c}{\textbf{(c) Belief-shaping success rate}} \\
\cmidrule(lr){2-4}\cmidrule(lr){5-8}\cmidrule(l){9-11}
\textbf{Model}
& \textbf{Align Acc} & \textbf{Wolf F1} & \textbf{Role Pred}
& \textbf{Guard} & \textbf{Seer} & \textbf{Villager} & \textbf{Witch}
& \textbf{GPT-4o-mini} & \textbf{GPT-4o} & \textbf{Claude-4.5} \\
\midrule
Base  & $40.2_{\pm 4.3}$ & $39.8_{\pm 4.1}$ & $37.8_{\pm 4.6}$
      & 43.2 & 43.4 & 27.2 & 37.4
      & 82.2 & 33.4 & 70.6 \\
ReCon & $38.2_{\pm 4.7}$ & $45.2_{\pm 4.2}$ & $37.4_{\pm 4.5}$
      & 31.6 & 46.8 & 41.6 & 29.6
      & 78.8 & 36.4 & 44.2 \\
MaKTO & $44.8_{\pm 4.6}$ & $31.2_{\pm 4.8}$ & $43.2_{\pm 4.4}$
      & 43.4 & 55.4 & 24.2 & 49.8
      & 75.5 & 36.4 & 42.2 \\
Ours  & $\mathbf{66.0_{\pm 3.2}}$ & $\mathbf{56.4_{\pm 2.5}}$ & $\mathbf{57.2_{\pm 3.5}}$
      & \textbf{52.4} & \textbf{66.4} & \textbf{52.2} & \textbf{57.8}
      & \textbf{88.2} & \textbf{78.4} & \textbf{87.8} \\
\bottomrule
\end{tabular}%
}
\end{table*}

\begin{table*}[!t]
    \centering
    \begin{minipage}[t]{0.48\textwidth}
        \centering
        \includegraphics[
            width=\linewidth,
            height=0.467\linewidth
        ]{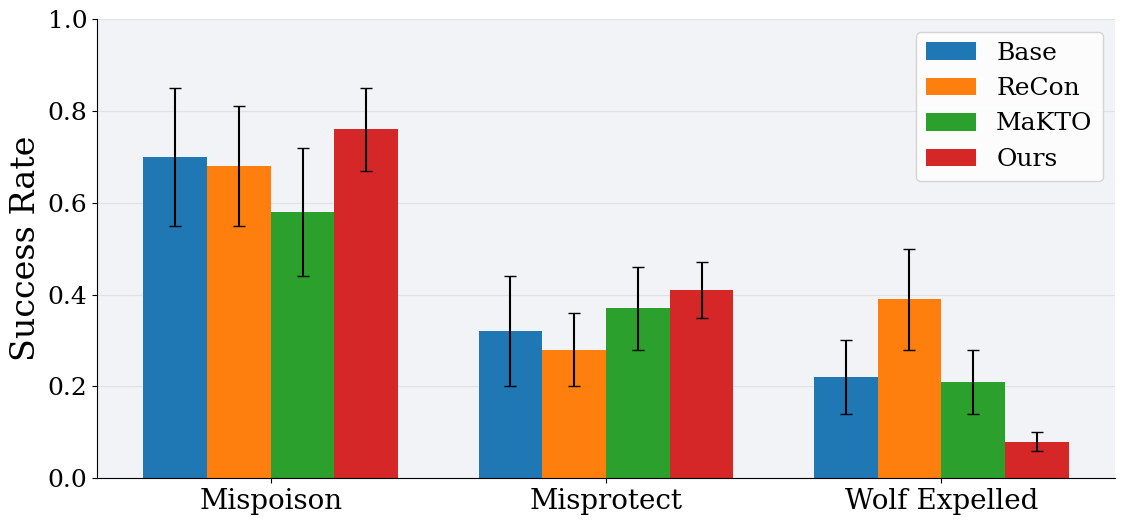}
        \captionsetup{skip=0.3em}
        \captionof{figure}{Metrics for Werewolf-side belief shaping, measuring whether Werewolves mislead the Villager-side.}
        \label{fig:night_skill_wolf}
    \end{minipage}
    \hfill
    \begin{minipage}[t]{0.48\textwidth}
        \centering
        \includegraphics[
            width=\linewidth,
            height=0.467\linewidth
        ]{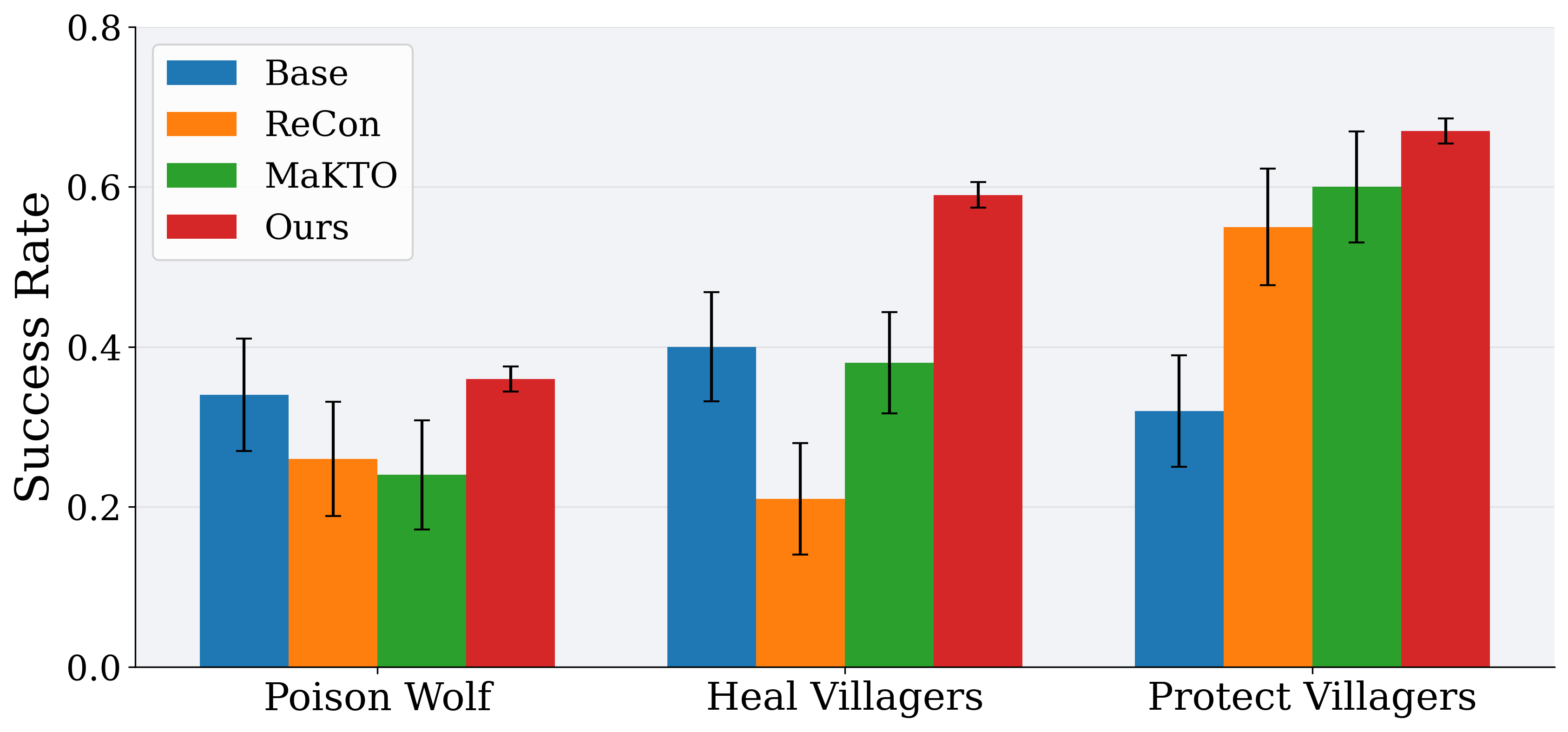}
        \captionsetup{skip=0.3em}
        \captionof{figure}{Belief-grounded night-action target selection for special roles.}
        \label{fig:night_action_accuracy}
    \end{minipage}
\end{table*}

\begin{table*}[!t]
\centering
\begin{minipage}[t]{0.48\textwidth}
\vspace{0pt}
\centering
\small
\color{black}
\arrayrulecolor{black}
\captionsetup{
    skip=2pt,
    labelfont={color=black},
    textfont={color=black}
}
\captionof{table}{Effect of belief quality on role-prediction accuracy and game performance (win rate).}
\label{tab:belief_quality}
\setlength{\tabcolsep}{3pt}
\renewcommand{\arraystretch}{1.25}
\begin{tabularx}{\linewidth}{@{}l*{3}{>{\centering\arraybackslash}X}@{}}
\toprule
\textbf{Metric}
& \textbf{\makecell{MARBO\\(Original)}}
& \textbf{\makecell{MARBO\\(Random)}}
& \textbf{\makecell{MARBO\\(Oracle)}} \\
\midrule
Accuracy & $57.2_{\pm 3.5}$ & $18.3_{\pm 2.7}$ & $100.0$ \\
Performance & $70.2_{\pm 2.6}$ & $30.1_{\pm 3.6}$ & $78.5_{\pm 2.9}$ \\
\bottomrule
\end{tabularx}
\arrayrulecolor{black}
\end{minipage}
\hfill
\begin{minipage}[t]{0.48\textwidth}
\vspace{0pt}
\centering
\small
\color{black}
\arrayrulecolor{black}
\captionsetup{skip=2pt}
\captionof{table}{Belief-grounded task action performance (\%).}
\label{tab:belief_to_vote_action}
\setlength{\tabcolsep}{3pt}
\renewcommand{\arraystretch}{1.12}
\begin{tabularx}{\linewidth}{@{}l*{3}{>{\centering\arraybackslash}X}@{}}
\toprule
\textbf{Model}
& \textbf{Wolf F1}
& \textbf{Vote Acc}
& \textbf{Expel Wolf} \\
\midrule
Base & 39.8 & 51.2 & 58.4 \\
ReCon & 45.2 & 46.0 & 40.4 \\
MaKTO & 31.2 & 64.0 & 52.2 \\
Ours & \textbf{56.4} & \textbf{70.2} & \textbf{65.4} \\
\bottomrule
\end{tabularx}
\end{minipage}
\end{table*}
Table~\ref{tab:belief_all}(a) shows that MARBO infers hidden roles most accurately overall ($57\%$), and the role-wise breakdown in Table~\ref{tab:belief_all}(b) confirms that this gain is consistent rather than concentrated in a single role. 
The largest improvement appears for the Villager role, whose identity must be inferred from discussion alone, indicating that MARBO stays reliable without private evidence. 
These results suggest MARBO forms well-calibrated relational beliefs, reducing both false and missed accusations.

Figure~\ref{fig:long_horizon_reasoning} shows that these gains persist through Day 4, indicating that explicit relational belief supervision provides stable and temporally coherent belief estimates.

\textcolor{black}{As shown in Table~\ref{tab:belief_quality}, random beliefs achieve near-chance role prediction $(18.3\%)$ and substantially degrade game performance to $30.1\%$. Oracle beliefs achieve perfect role prediction and improve performance to $78.5\%$, although winning still depends on subsequent reasoning and strategic decisions.}
\textcolor{black}{MARBO attains $57.2\%$ belief accuracy, substantially above chance, and narrows much of the performance gap toward the oracle. These results better illustrate why accurate social-label inference matters and provide context for interpreting MARBO's belief accuracy.}

\paragraph{Belief-shaping communication optimization.}
\textcolor{black}{For \textbf{Q3}, we assess MARBO's ability to shape other agents' beliefs through communication.} Figure~\ref{fig:night_skill_wolf} shows that MARBO-trained Werewolf-side agents induce more Villager-side errors: increasing Mispoison by \texttt{witch} and Misprotect by \texttt{guard} while reducing Wolf Expelled \textcolor{black}{indicating that MARBO better manipulates others' beliefs while concealing its own werewolf identity.}

Table~\ref{tab:belief_all}(c) further shows that opponents more often infer the speaker as the intended role, such as perceiving a Werewolf as the Seer after fake-Seer speech \textcolor{black}{indicating that successful misdirection arises from MARBO's ability to manipulate opponents' beliefs.} 
\textcolor{black}{More details about the experiments are provided in Appendix~\ref{subsecapp:belief_shaping_experiments_details}.}

\paragraph{Belief-grounded task action optimization.}

To answer \textbf{Q4}, we evaluate whether MARBO uses relational beliefs to optimize target-directed task actions in Werewolf. Table~\ref{tab:belief_to_vote_action} evaluates the belief-to-vote pipeline with three metrics. Werewolf F1-score measures role-belief accuracy, \textcolor{black}{reflecting how precisely an agent identifies werewolves without over-accusing villagers.} Vote Accuracy measures correct target selection, \textcolor{black}{indicating how often an agent votes for actual werewolf.} Expel Wolf measures the voting outcome, \textcolor{black}{capturing whether these individual votes converge into a correct collective decision.} MARBO achieves the best scores on all metrics ($56\%$, $70\%$, and $65\%$), indicating that belief-grounded feedback improves the path from target belief formation to task outcomes.

Figure~\ref{fig:night_action_accuracy} extends this analysis to night actions, where special roles must commit to a single target without any public discussion. \textcolor{black}{Poison Wolf and Heal Villagers measure whether the Witch uses its one-shot potions on appropriate targets, and Protect Villagers measures whether the Guard shields a villager who is actually targeted that night.} MARBO leads on all three, demonstrating that its relational beliefs also guide private, irreversible decisions.

\subsection{Ablation Studies}
\label{sec:ablation}

\begin{table}[t]
\centering
\captionsetup{skip=2pt}
\small
\setlength{\tabcolsep}{4pt}
\renewcommand{\arraystretch}{1.12}

\caption{
Ablation study of MARBO with Qwen2.5-7B.
}
\label{tab:ablation_marbo}

\begin{tabular}{@{}lc@{}}
\toprule

\textbf{Variant} & \textbf{Win Rate (\%)} \\

\midrule

\textbf{MARBO} & $\mathbf{65.6_{\pm 1.7}}$ \\

w/o Belief construction & $55.0_{\pm 2.6}$ \\

w/o Belief-Action optimization & $39.0_{\pm 3.4}$ \\

w/o Belief-Shaping optimization & $34.0_{\pm 3.8}$ \\

\bottomrule
\end{tabular}
\end{table}

We quantify the contribution of each MARBO component in Table~\ref{tab:ablation_marbo}.
Removing belief grounding lowers the average win rate from $66\%$ to $55\%$, while removing belief-action feedback drops it to $39\%$, showing that relational beliefs must be both constructed and connected to target-directed decisions. The largest degradation occurs without belief-speech feedback, reducing the win rate to $34\%$ and highlighting the importance of belief-shaping communication. These results show that MARBO's gains arise from jointly optimizing relational beliefs, task actions, and speech actions.

\section{Conclusion}
\vspace{-0.5em}

We propose MARBO, a relational-belief-driven preference optimization framework for LLM agents in partially observable social deduction games. MARBO supervises agent-wise beliefs about hidden roles, team alignments, and strategic relations, then constructs belief-grounded feedback for task decisions and speech actions. This aligns behavior with inferred social relations and encourages communication that shapes others' beliefs. Across \textit{Werewolf} and \textit{Among Us}, MARBO outperforms prompting-based and preference-learning baselines, demonstrating the value of optimizing hidden social reasoning in partially observable multi-agent interaction.

\clearpage
\section*{Limitations} 
Our framework has several limitations to address in future research. 
First, our framework is limited to turn-based interactions, restricting communication to predefined turns.
Second, our model may occasionally hallucinate, resulting in actions inconsistent with the game trajectory. 
Finally, offline training restricts exploration to strategies observed in the training data.
Future work includes enabling online learning for exploration and extension to unrestricted communication.

\section*{Ethical Considerations}
Our study investigates deception within SDGs, an inherent part of gameplay. 
Nevertheless, our findings may carry dual-use risks, as the learned behaviors could potentially be exploited to improve deceptive capabilities.
We strongly oppose the use of our work for deceptive or manipulative practices. 
We therefore advocate appropriate safeguards and ethical guidelines to mitigate potential misuse.

\section*{Artifact Licenses and Intended Use}
We rely only on SDG benchmarks, including Among Us and Werewolf, released for research use. 
Our use remains within each benchmark’s original research intent.
All models and datasets used in this work are publicly available and used in accordance with their respective licenses and intended use.

\section*{Acknowledgment}
This work was supported partly by the National Research Foundation of Korea (NRF) grant funded by the Korea government (MSIT) 
(No. RS-2025-23523191, LLM-Based Multi-Agent Reinforcement Learning for End-to-End Large AutonomousSwarm Control),
the Leading Generative AI Human Resources Development(RS-2025-25441313) grant funded by the Korea government(MSIT), 
the Institute of Information \& Communications Technology Planning \& Evaluation (IITP) grant funded by the Korea government (MSIT)
(No. RS-2022-II220469, Development of Core Technologies for Task-oriented Reinforcement Learning for Commercialization of Autonomous Drones),
(No. RS-2025-25442824, AI Star-Fellowship Program (UNIST)),
and (No. RS-2020II201336, Artificial Intelligence Graduate School Support(UNIST)).


\bibliography{custom}

\clearpage

\appendix
\twocolumn

\renewcommand{\theequation}{A.\arabic{equation}}
\renewcommand{\theHequation}{A.\arabic{equation}}
\setcounter{equation}{0}
\renewcommand{\thefigure}{A.\arabic{figure}}
\renewcommand{\theHfigure}{A.\arabic{figure}}
\setcounter{figure}{0}
\renewcommand{\thetable}{A.\arabic{table}}
\renewcommand{\theHtable}{A.\arabic{table}}
\setcounter{table}{0}

\section{Details of Social Deduction Games}
\label{secapp:social_deduction_games_details}
This section describes the details of two prominent social deduction games used in our work: the \textit{Werewolf} game and the \textit{Among Us} game.
Section~\ref{subsecapp:werewolf_details} covers the objectives, game process, and agent roles of the Werewolf game, while Section~\ref{subsecapp:amongus_details} presents the corresponding details for the Among Us game.

\subsection{Werewolf Game Details}
\label{subsecapp:werewolf_details}

\paragraph{Game Rules.}\mbox{}\\  
Werewolf (also known as Mafia) is a classic social deduction game in which players are secretly assigned to one of two competing factions: the \emph{Werewolf} faction or the \emph{Villager} faction.
The game alternates between a Night Action Phase, during which special roles perform hidden actions, and a Day Phase, during which all surviving players engage in open discussion and cast a public vote to eliminate a suspect.
A key asymmetry drives the game: the Werewolf faction possesses full knowledge of its own members' identities, whereas Villager-faction players must infer hidden allegiances solely from observed behavior and spoken claims.
This information asymmetry forces Villager-faction players to engage in deductive reasoning and persuasion, while Werewolf-faction players must craft credible deceptions to conceal their true identities.
 
In our experimental setup, we adopt a 9-player \emph{Seer-Witch-Guard} configuration consisting of three Werewolves, one Seer, one Witch, one Guard, and three ordinary Villagers.
Each player is assigned a unique player index (P1–P9) and a role kept secret from all other players except fellow Werewolves.

\begin{figure*}[!ht]
    \vspace{-0.5em}
    \centering
    \includegraphics[width=0.98\textwidth]{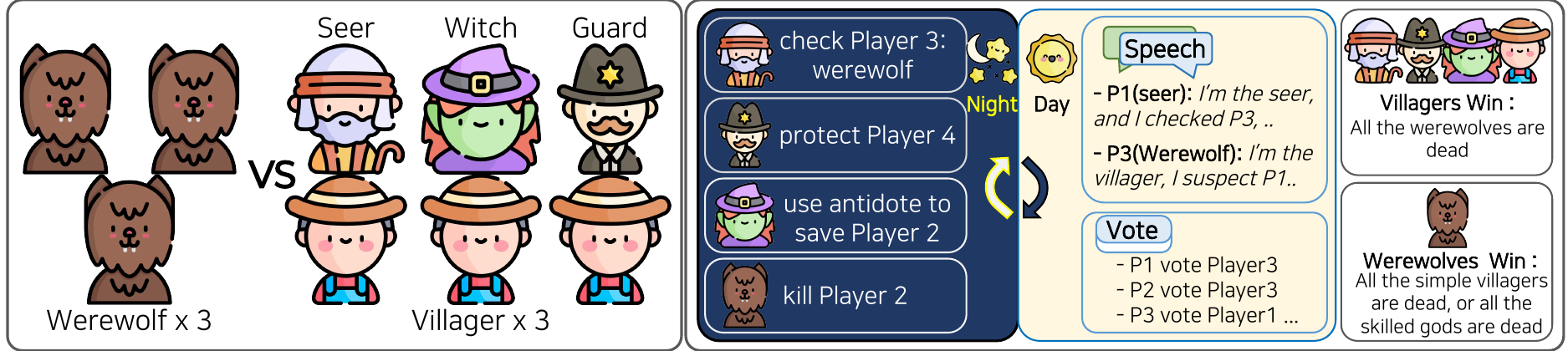}
    \caption{Illustration of 9-player Werewolf game configuration with roles of Werewolf, Seer, Witch, Guard and Villagers and gameplay process of alteration of night and day.}
    \label{fig:app_werewolf}
    \vspace{-1em}
\end{figure*}

\paragraph{Game Objectives.}\mbox{}\\
The two factions pursue diametrically opposed objectives:
\begin{itemize}
    \item \textbf{Werewolf Faction.} The Werewolves win if, at the start of any Day Phase, the number of surviving Werewolves equals or exceeds the number of surviving Villager-faction players (i.e., Werewolves successfully outnumber the remaining opposition).
    \item \textbf{Villager Faction.} The Villagers, Seer, Witch, and Guard collectively win if all Werewolves are eliminated through nighttime actions or daytime voting before the Werewolves achieve numerical dominance.
\end{itemize}
Because each player can speak freely and may claim any role, strategic deception and persuasion are central mechanisms through which both factions pursue victory.

\paragraph{Game Process.}\mbox{}\\
A complete game of Werewolf proceeds through the following alternating phases:
\begin{itemize}
    \item \textbf{Night Action Phase :}
    At the beginning of each night, all players ``close their eyes'' (i.e., their observations are restricted to their own role's actions).
    Special roles then act sequentially and privately. 
    At the end of the Night Action Phase, the game moderator announces which players (if any) have been eliminated. After the Night Action Phase, all surviving players participate in the Day Phase, which consists of two stages: Speech Phase and Voting Phase.

    \item \textbf{Speech Phase :} Each player delivers a speech in a fixed sequential order (P1 through P$N$, skipping eliminated players). Players may claim any role, share deductions, accuse others, or defend themselves. Werewolves typically impersonate Villager-faction roles (especially the Seer) to mislead the group, while Villager-faction players attempt to identify inconsistencies in others' statements.

    \item \textbf{Voting Phase :} After all speeches, each surviving player publicly votes to eliminate one other player. The player who receives the most votes is eliminated; in case of a tie, no elimination occurs. After the vote, it is revealed whether the eliminated player (if any) was a Werewolf or a Villager-faction member.
\end{itemize}

\begin{figure*}[!t]
    \vspace{-0.5em}
    \centering
    \includegraphics[width=0.98\textwidth]{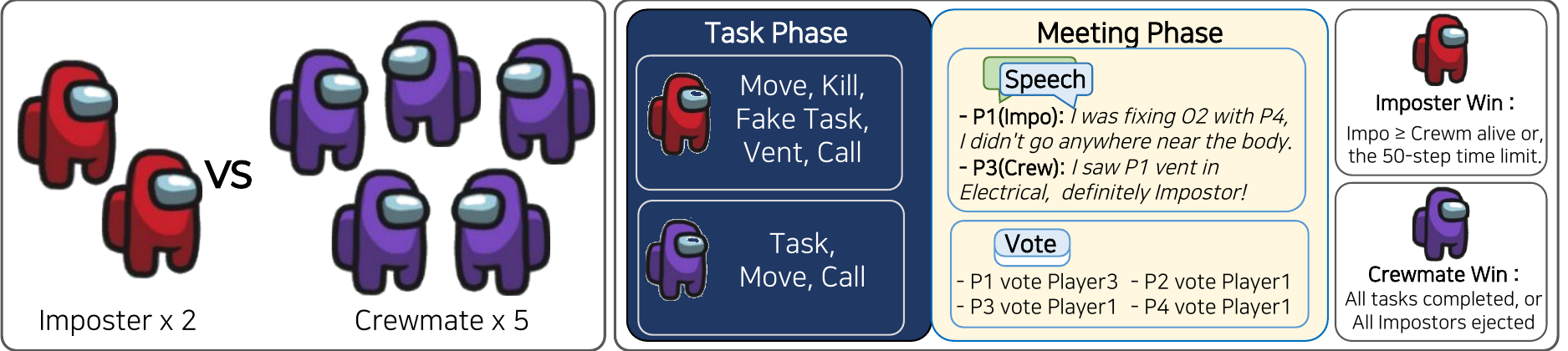}
    \caption{Illustration of Among Us game configuration with roles of Imposters and Crewmates and gameplay process of alteration of task phase and meeting phase.}
    \label{fig:app_amongus}
    \vspace{-1em}
\end{figure*}

\paragraph{Role Descriptions.}\mbox{}\\
The 9-player Werewolf game has five different roles: Werewolf, Villager, Seer, Witch, Guard, as following abilities;
\begin{itemize}
    \item \textbf{Werewolf (×3)} knows all fellow Werewolves; each night collectively eliminates one Villager-faction player and must conceal their identity through deceptive speech during the day.
    \item \textbf{Villager (×3)} has no special nighttime ability; contributes solely through daytime discussion and voting to help identify and eliminate Werewolves.
    \item \textbf{Seer (×1)} investigates one player and learns whether they are a Werewolf each night privately; must decide strategically when and how to reveal this information during the day.
    \item \textbf{Witch (×1)} holds a one-time antidote to save the night's victim and a one-time poison to eliminate any player; cannot use both in the same night.
    \item \textbf{Guard (×1)} protects one player from Werewolf elimination each night; cannot protect the same player on two consecutive nights.
\end{itemize}


\subsection{Among Us Game Details}
\label{subsecapp:amongus_details}

\paragraph{Game Rules.}\mbox{}\\
Among Us is a social deduction and deception game adapted as a text-based multi-agent sandbox for LLM agents.
Players are secretly assigned to one of two roles: \emph{Crewmate} or \emph{Impostor}.
Unlike Werewolf, Among Us incorporates spatial navigation across a map and task execution alongside discussion and voting, requiring Impostors to mimic Crewmate behavior while concealing kills.
Each game instance begins with 5 Crewmates and 2 Impostors, and players act sequentially at each time step.

\paragraph{Game Objectives.}\mbox{}
\begin{itemize}
    \item \textbf{Crewmates} win by (a) completing all assigned tasks, or (b) voting out all Impostors.
    \item \textbf{Impostors} win by (a) reducing Crewmates until $|\text{Impostors}| \geq |\text{Crewmates}|$, or (b) exhausting the 50-step time limit.
\end{itemize}
Fake tasks performed by Impostors do not contribute to task completion progress.

\renewcommand{\thetable}{B.\arabic{table}}
\renewcommand{\theHtable}{B.\arabic{table}}
\setcounter{table}{0}

\begin{table*}[!t]
\centering
\small
\setlength{\tabcolsep}{5pt}
\renewcommand{\arraystretch}{1.25}
\caption{Role-conditional construction of belief supervision labels in Werewolf}
\label{tab:belief_label_construction}
\begin{tabularx}{\textwidth}{
>{\raggedright\arraybackslash}p{2.0cm}
>{\raggedright\arraybackslash}X
>{\raggedright\arraybackslash}X
}
\toprule
\textbf{Role of agent \(i\)}
& \textbf{Before Day-1 speech}
& \textbf{After Day-1 speech} \\
\midrule

\textbf{Werewolf}
&
\begin{itemize}[leftmargin=1em,topsep=0pt,itemsep=0pt]
    \item Fellow werewolves: \texttt{Werewolf}
    \item All other agents: \texttt{Unknown}
\end{itemize}
&
\begin{itemize}[leftmargin=1em,topsep=0pt,itemsep=0pt]
    \item Ground-truth social label
\end{itemize}
\\
\midrule

\textbf{Seer}
&
\begin{itemize}[leftmargin=1em,topsep=0pt,itemsep=0pt]
    \item Checked player is \texttt{Werewolf}: \texttt{Werewolf}
    \item Checked player is not \texttt{Werewolf}: \texttt{Villager-side}
    \item All other agents: \texttt{Unknown}
\end{itemize}
&
\begin{itemize}[leftmargin=1em,topsep=0pt,itemsep=0pt]
    \item Ground-truth social label
\end{itemize}
\\
\midrule

\textbf{Others}
&
\begin{itemize}[leftmargin=1em,topsep=0pt,itemsep=0pt]
    \item All other agents: \texttt{Unknown}
\end{itemize}
&
\begin{itemize}[leftmargin=1em,topsep=0pt,itemsep=0pt]
    \item Ground-truth social label
\end{itemize}
\\
\bottomrule
\end{tabularx}
\end{table*}

\begin{table*}[!t]
\centering
\small
\setlength{\tabcolsep}{5pt}
\renewcommand{\arraystretch}{1.25}
\caption{Role-conditional construction of belief-supervision labels in Among Us.}
\label{tab:belief_label_construction_amongus}
\begin{tabularx}{\textwidth}{
>{\raggedright\arraybackslash}p{2.0cm}
>{\raggedright\arraybackslash}X
>{\raggedright\arraybackslash}X
}
\toprule
\textbf{Role of agent \(i\)}
& \textbf{Before two discussion rounds}
& \textbf{After two discussion rounds} \\
\midrule

\textbf{Impostor}
&
\begin{itemize}[leftmargin=1em,topsep=0pt,itemsep=0pt]
    \item Ground-truth social label
\end{itemize}
&
\begin{itemize}[leftmargin=1em,topsep=0pt,itemsep=0pt]
    \item Ground-truth social label
\end{itemize}
\\
\midrule

\textbf{Crewmate}
&
\begin{itemize}[leftmargin=1em,topsep=0pt,itemsep=0pt]
    \item All other agents: \texttt{Unknown}
\end{itemize}
&
\begin{itemize}[leftmargin=1em,topsep=0pt,itemsep=0pt]
    \item Ground-truth social label
\end{itemize}
\\

\bottomrule
\end{tabularx}
\end{table*}

\begin{table*}[!t]
\centering
\small
\caption{Role-specific outcome conditions for speech-action preference.}
\label{tab:speech_outcome_conditions}
\begin{tabular}{p{0.10\linewidth} p{0.40\linewidth} p{0.43\linewidth}}
\toprule
Game & Favorable outcomes & Unfavorable outcomes \\
\midrule
\textit{Werewolf}
&
\begin{itemize}[leftmargin=*,topsep=0pt,itemsep=0pt]
    \item \texttt{Werewolf} avoids expulsion
    \item \texttt{Villager} receives no votes
    \item \texttt{Seer} receives no \texttt{Villager} votes
\end{itemize}
&
\begin{itemize}[leftmargin=*,topsep=0pt,itemsep=0pt]
    \item \texttt{Werewolf} is expelled or receives majority \texttt{Villager} votes
    \item \texttt{Villager} is expelled
    \item \texttt{Witch} receives \texttt{Werewolf} and multiple \texttt{Villager} votes
    \item \texttt{Seer} receives majority \texttt{Villager} votes
\end{itemize}
\\
\midrule
\textit{Among Us}
&
\begin{itemize}[leftmargin=*,topsep=0pt,itemsep=0pt]
    \item \texttt{Crewmate} speaker receives no votes
    \item \texttt{Impostor} avoids majority votes or ejection
\end{itemize}
&
\begin{itemize}[leftmargin=*,topsep=0pt,itemsep=0pt]
    \item \texttt{Crewmate} speaker receives majority votes
    \item \texttt{Impostor} speaker is ejected or receives majority votes
\end{itemize}
\\
\bottomrule
\end{tabular}
\vspace{0.6em}
\caption{Favorable belief transitions for speech-action preference.}
\label{tab:speech_belief_shift_conditions}
\begin{tabular}{p{0.10\linewidth} p{0.25\linewidth} p{0.55\linewidth}}
\toprule
Game & Speaker role & Favorable belief transition for listener \(j\)'s belief about speaker \(i\) \\
\midrule
\multirow{2}{*}{\textit{Werewolf}}
&
\texttt{Werewolf}
&
\[
\hat b_t^{ji}\in\{\texttt{Werewolf}\}
\rightarrow
\hat b_{t,+}^{ji}\in\{\texttt{unknown},\texttt{Villager},\texttt{Seer},\texttt{Witch}\}
\]
\\
\cmidrule(lr){2-3}
&
\texttt{Villager}, \texttt{Seer}, \texttt{Witch}
&
\[
\hat b_t^{ji}\in\{\texttt{Werewolf}\}
\rightarrow
\hat b_{t,+}^{ji}\in\{\texttt{unknown},\texttt{Villager},\texttt{Seer},\texttt{Witch}\}
\]
\\
\midrule
\multirow{2}{*}{\textit{Among Us}}
&
\texttt{Impostor}
&
\[
\hat b_t^{ji}\in\{\texttt{Impostor}\}
\rightarrow
\hat b_{t,+}^{ji}\in\{\texttt{unknown},\texttt{Crewmate}\}
\]
\\
\cmidrule(lr){2-3}
&
\texttt{Crewmate}
&
\[
\hat b_t^{ji}\in\{\texttt{Impostor}\}
\rightarrow
\hat b_{t,+}^{ji}\in\{\texttt{unknown},\texttt{Crewmate}\}
\]
\\
\bottomrule
\end{tabular}
\vspace{-0.8em}
\end{table*}

\begin{table*}[!t]
\centering
\small
\setlength{\tabcolsep}{5pt}
\renewcommand{\arraystretch}{1.12}
\setlist[itemize]{leftmargin=1.1em, nosep}
\caption{Step-level preference selection criteria for Werewolf game actions.}
\label{tab:selection_methods_werewolf}
\begin{tabularx}{\textwidth}{
>{\raggedright\arraybackslash}p{1.15cm}
>{\raggedright\arraybackslash}p{1.15cm}
>{\raggedright\arraybackslash}p{1.25cm}
>{\raggedright\arraybackslash}X
>{\raggedright\arraybackslash}X
}
\toprule
\textbf{Output} & \textbf{Stage} & \textbf{Method} & \textbf{Desirable} & \textbf{Undesirable} \\
\midrule

\multirow{2}{*}{\textbf{Action}}
& \textbf{Night}
& \textbf{Heuristic}
&
\begin{itemize}
    \item Werewolf kills Seer
    \item Witch poisons Werewolf
    \item Witch heals
    \item Werewolf is poisoned after Day 1
\end{itemize}
&
\begin{itemize}
    \item No Werewolf kill
    \item No Witch healing on Night 1
    \item Villager is poisoned
    \item Werewolf is protected
\end{itemize}
\\
\cmidrule(lr){2-5}

& \textbf{Vote}
& \textbf{Heuristic}
&
\begin{itemize}
    \item Villager-side players vote for a Werewolf
\end{itemize}
&
\begin{itemize}
    \item Villager-side players vote for a Villager
    \item Abstention
    \item Vote split away from the true Seer
    \item Vote split away from the majority when no Seer is present
\end{itemize}
\\

\bottomrule
\end{tabularx}
\vspace{0.6em}
\setlength{\tabcolsep}{5pt}
\renewcommand{\arraystretch}{1.12}
\setlist[itemize]{leftmargin=1.1em, nosep}
\caption{Step-level preference selection criteria for Among Us game actions.}
\label{tab:selection_methods_amongus}
\begin{tabularx}{\textwidth}{
>{\raggedright\arraybackslash}p{1.15cm}
>{\raggedright\arraybackslash}p{1.15cm}
>{\raggedright\arraybackslash}p{1.25cm}
>{\raggedright\arraybackslash}X
>{\raggedright\arraybackslash}X
}
\toprule
\textbf{Output} & \textbf{Stage} & \textbf{Method} & \textbf{Desirable} & \textbf{Undesirable} \\
\midrule

\multirow{2}{*}{\textbf{Action}}
& \textbf{Task}
& \textbf{Heuristic}
&
\begin{itemize}
    \item Crewmate completes a task
    \item Crewmate reports a dead body
    \item Crewmate calls a meeting with hard evidence
    \item Impostor kills a Crewmate with no witnesses
    \item Impostor moves or vents after their own kill
\end{itemize}
&
\begin{itemize}
    \item Crewmate calls a meeting without evidence
    \item Crewmate ignores a visible dead body
    \item Crewmate repeatedly ignores an available task
    \item Impostor kills with Crewmate witnesses
    \item Invalid action under the game rules
\end{itemize}
\\
\cmidrule(lr){2-5}

& \textbf{Vote}
& \textbf{Heuristic}
&
\begin{itemize}
    \item Crewmate votes for an Impostor
\end{itemize}
&
\begin{itemize}
    \item Crewmate votes for a Crewmate
\end{itemize}
\\

\bottomrule
\end{tabularx}
\end{table*}

\paragraph{Game Process.}\mbox{}\\
A complete game of Among Us proceeds through the following alternating phases:
\begin{itemize}
    \item \textbf{Task Phase :} The game begins in the Task Phase and resumes after each inconclusive meeting. Crewmates complete assigned tasks at designated map locations, while Impostors blend in by performing fake tasks, killing isolated Crewmates (subject to a kill cooldown), and using ventilation shafts to reposition undetected.
    Discovering a dead body immediately triggers a Meeting Phase.

    \item \textbf{Meeting Phase :} Triggered by a reported body or an emergency button press, the Meeting Phase convenes all surviving players for 3 rounds of open discussion, followed by a simultaneous vote to eject a suspect. The player with the most votes is eliminated; if no majority is reached, the game returns to the Task Phase. The game terminates when any win condition is satisfied.
\end{itemize}

\paragraph{Role Descriptions.}\mbox{}\\
The Among Us game features two roles, Crewmate and Impostor, with the following abilities:
\begin{itemize}
    \item \textbf{Crewmate (×5)} completes assigned tasks at specific map locations and uses discussion and voting to identify and eject Impostors; available actions are \{\textsc{Move}, \textsc{Complete Task}, \textsc{Speak}, \textsc{Vote}, \textsc{Report Body}, \textsc{Call Meeting}, \textsc{Check Security Camera}\}.
    \item \textbf{Impostor (×2)} wins by eliminating Crewmates or stalling time; can perform fake tasks and uniquely has access to \textsc{Kill} (eliminates an adjacent Crewmate, subject to cooldown) and \textsc{Vent} (instant travel between ventilation shafts) in addition to all Crewmate actions except \textsc{Complete Task}.
\end{itemize}

\newcommand{\experimentalDetailsAppendix}{%
\clearpage
\onecolumn

\renewcommand{\theequation}{E.\arabic{equation}}
\renewcommand{\theHequation}{E.\arabic{equation}}
\setcounter{equation}{0}
\renewcommand{\thefigure}{E.\arabic{figure}}
\renewcommand{\theHfigure}{E.\arabic{figure}}
\setcounter{figure}{0}
\renewcommand{\thetable}{E.\arabic{table}}
\renewcommand{\theHtable}{E.\arabic{table}}
\setcounter{table}{0}

\section{Experimental Details}
\label{app:experimental_details}

This section provides the metric definitions and measurement procedures for relational belief construction in appendix~\ref{subsecapp:relational_belief_experiments_details}, belief-shaping communication in appendix~\ref{subsecapp:belief_shaping_experiments_details}, and belief-grounded action selection in appendix~\ref{subsecapp:belief_action_experiments_details}.

\subsection{Relational Belief Construction}
\label{subsecapp:relational_belief_experiments_details}

We describe how to compute the game-specific metrics used to evaluate relational belief construction in \textit{Werewolf} and \textit{Among Us}, including exact role prediction, ally-opponent alignment, and adversary identification accounting for missed targets and false accusations.

\paragraph{Werewolf}

\begin{itemize}[leftmargin=*, itemsep=0.5em, topsep=0.2em]
\item \textbf{Alignment Accuracy (Align Acc).}
Alignment Accuracy measures binary ally-adversary judgments as follows:
\begin{equation*}
\text{Align Acc} = \frac{\#\,\text{correct alignments}}{\#\,\text{agents}}
\end{equation*}

\item \textbf{Role Prediction Accuracy (Role Pred).}
Role Prediction Accuracy measures exact hidden-role inference, computed as:
\begin{equation*}
\text{Role Pred} = \frac{\#\,\text{correct roles}}{\#\,\text{agents}}
\end{equation*}

\item \textbf{Wolf F1.}
Wolf F1 measures werewolf identification while accounting for both false accusations and missed wolves, computed as:
\begin{equation*}
\text{Wolf F1} = \frac{2\,\#\,\text{wolves found}}
{2\,\#\,\text{wolves found} + \#\,\text{false accusations} + \#\,\text{wolves missed}}
\end{equation*}
\end{itemize}

\paragraph{Among Us}

\begin{itemize}[leftmargin=*, itemsep=0.5em, topsep=0.2em]
\item \textbf{Role Accuracy (Role Acc.)}
Role Accuracy measures exact role identification and is computed as follows:
\begin{equation*}
\text{Role Acc.} =
\frac{\#\{\text{correct role predictions}\}}
{\#\{\text{role predictions}\}}
\end{equation*}

\item \textbf{Impostor Detection (Imp. Detect)}
Impostor Detection measures true Impostor identification and is computed as follows:
\begin{equation*}
\text{Imp. Detect} =
\frac{\#\{\text{Impostors correctly identified}\}}
{\#\{\text{true Impostors}\}}
\end{equation*}

\item \textbf{Belief F1}
Belief F1 captures balanced Impostor identification by accounting for both false accusations and missed Impostors, computed as:
\begin{equation*}
\text{Belief F1} =
\frac{2 \cdot \text{Precision}_{\mathrm{Imp}}
      \cdot \text{Recall}_{\mathrm{Imp}}}
{\text{Precision}_{\mathrm{Imp}} + \text{Recall}_{\mathrm{Imp}}}
\end{equation*}
\end{itemize}

\clearpage
\subsection{Belief-shaping Communication}
\label{subsecapp:belief_shaping_experiments_details}

We detail the game-specific measures for belief-shaping communication in \textit{Werewolf} and \textit{Among Us}, capturing whether speech steers others' beliefs and leads to favorable individual or collective outcomes.

\paragraph{Werewolf}

\begin{itemize}[leftmargin=*, itemsep=0.5em, topsep=0.2em]
\item \textbf{Belief-shaping Success}
We evaluate whether an agent's speech shapes other agents' beliefs toward its intended identity. 
For each speaker $i$, we reconstruct each listener $j$'s perspective using the interaction history and the speaker's utterance, and apply the CoT-style evaluator in Appendix~\ref{app:prompt} to infer how $j$ perceives $i$'s role. 
For example, if a Werewolf delivers fake-Seer speech, the intended identity is Seer. 
A trial is successful when the evaluator predicts the speaker as this intended role, and we average the success rate across listeners, speeches, and games.

\item \textbf{MisPoison}
MisPoison measures how often a Witch incorrectly poisons a non-werewolf and is computed as the proportion of such targets among all poison actions:
\begin{equation*}
\text{MisPoison} = \frac{\#\{\text{non-wolves poisoned}\}}{\#\{\text{poison actions}\}}
\end{equation*}

\item \textbf{MisProtect}
MisProtect measures how often a Guard protects a werewolf and is computed as the proportion of werewolf targets among all protect actions:
\begin{equation*}
\text{MisProtect} = \frac{\#\{\text{werewolves protected}\}}{\#\{\text{protect actions}\}}
\end{equation*}

\item \textbf{Wolf Expelled}
Wolf Expelled measures how often collective voting results in the elimination of a werewolf and is computed as the proportion of werewolves among all expelled players:
\begin{equation*}
\text{Wolf Expelled} = \frac{\#\{\text{werewolves expelled}\}}{\#\{\text{expelled players}\}}
\end{equation*}
\end{itemize}

\paragraph{Among Us}

\begin{itemize}[leftmargin=*, itemsep=0.5em, topsep=0.2em]
\item \textbf{Suspicion Reduced}
Suspicion Reduced measures how often Crewmates lower their suspicion toward an Impostor speaker after its speech and is computed as the proportion of reduced-suspicion updates:
\begin{equation*}
\text{Suspicion Reduced} =
\frac{\#\{\text{Crewmate beliefs with reduced suspicion}\}}
{\#\{\text{updated Crewmate beliefs}\}}
\end{equation*}

\item \textbf{Suspicion Raised}
Suspicion Raised measures how often Crewmates increase their suspicion toward an Impostor speaker after its speech and is computed as the proportion of increased-suspicion updates:
\begin{equation*}
\text{Suspicion Raised} =
\frac{\#\{\text{Crewmate beliefs with increased suspicion}\}}
{\#\{\text{updated Crewmate beliefs}\}}
\end{equation*}
\end{itemize}

\clearpage
\subsection{Belief-Grounded Action Selection}
\label{subsecapp:belief_action_experiments_details}

We describe how the game-specific metrics for belief-grounded decision making are measured in \textit{Werewolf} and \textit{Among Us}, including individual actions, voting, and group outcomes.

\paragraph{Werewolf}

\begin{itemize}[leftmargin=*, itemsep=0.5em, topsep=0.2em]
\item \textbf{Poison Wolf.}
Poison Wolf measures how often the Witch correctly targets a werewolf and is computed as the proportion of werewolves among all poison targets:

\begin{equation*}
\text{Poison Wolf} = \frac{\#\{\text{wolves poisoned}\}}{\#\{\text{poison actions}\}}
\end{equation*}

\item \textbf{MisPoison}
MisPoison measures how often the Witch incorrectly targets a non-werewolf and is computed as the proportion of non-werewolves among all poison targets:
\begin{equation*}
\text{MisPoison} = \frac{\#\{\text{non-wolves poisoned}\}}{\#\{\text{poison actions}\}}
\end{equation*}

\item \textbf{Protect Villager}
Protect Villager measures how often the Guard successfully protects a villager and is computed as the proportion of villagers among all protected targets:
\begin{equation*}
\text{Protect Villager} = \frac{\#\{\text{villagers protected}\}}{\#\{\text{protect actions}\}}
\end{equation*}

\item \textbf{Heal Villager}
Heal Villager measures how often the Witch successfully heals a villager and is computed as the proportion of villagers among all healed targets:
\begin{equation*}
\text{Heal Villager} = \frac{\#\{\text{villagers healed}\}}{\#\{\text{heal actions}\}}
\end{equation*}
\end{itemize}

\paragraph{Among Us}

\begin{itemize}[leftmargin=*, itemsep=0.5em, topsep=0.2em]
\item \textbf{Belief Vote}
Belief Vote measures whether an agent's vote is consistent with its inferred relational belief and is computed as the proportion of votes targeting players believed to be Impostors:
\begin{equation*}
\text{Belief Vote} =
\frac{\#\{\text{votes targeting believed Impostors}\}}
{\#\{\text{votes}\}}
\end{equation*}

\item \textbf{Vote Impostor (Vote Imp.)}
Vote Impostor measures how often an agent correctly votes for a true Impostor and is computed as the proportion of votes targeting true Impostors:
\begin{equation*}
\text{Vote Imp.} =
\frac{\#\{\text{votes targeting true Impostors}\}}
{\#\{\text{votes}\}}
\end{equation*}

\item \textbf{Expel Impostor (Expel Imp.)}
Expel Impostor measures how often collective voting ejects a true Impostor and is computed as the proportion of Impostors among all expelled players:
\begin{equation*}
\text{Expel Imp.} =
\frac{\#\{\text{Impostors expelled}\}}
{\#\{\text{players expelled}\}}
\end{equation*}
\end{itemize}
}


\renewcommand{\theequation}{B.\arabic{equation}}
\renewcommand{\theHequation}{B.\arabic{equation}}
\setcounter{equation}{0}
\renewcommand{\thefigure}{B.\arabic{figure}}
\renewcommand{\theHfigure}{B.\arabic{figure}}
\setcounter{figure}{0}

\section{Details of Belief-grounded Preference Design}
\label{app:reward_function}

This section presents the implementation details of our belief-grounded preference 
design.
Appendix~\ref{app:label_construction} describes the construction of belief-supervision labels.
Appendix~\ref{app:llm_judge} describes the LLM-judge protocol used to verify the preference labels assigned to generated speech samples.
Appendix~\ref{app:favorability} describes the criteria for strategic favorability in speech-action preference.
Appendix~\ref{app:task_utility} specifies the criteria that distinguish desirable from undesirable task samples.

\subsection{Belief-label Construction}
\label{app:label_construction}

\paragraph{Werewolf}
To address overconfident role prediction when little public evidence is available, we construct belief-supervision labels based on the information accessible to each role before and after Day-1 speech.
As shown in Table~\ref{tab:belief_label_construction}, before Day-1 speech, agents are supervised only on role information that is directly accessible from their role-specific observations.
Werewolves can identify fellow Werewolves, the Seer can label the checked player according to the inspection result, and other Villager-side agents assign \texttt{Unknown} to all other players.
All roles that cannot be reliably inferred from the agent's available information are assigned \texttt{Unknown} to prevent label leakage.
After Day-1 speech, labels are updated to ground-truth social labels, reflecting the information that becomes observable or reasonably inferable from the public discussion.

\paragraph{Among Us}
In Among Us, we construct belief-supervision labels based on whether two discussion rounds have been completed.
As shown in Table~\ref{tab:belief_label_construction_amongus}, before two discussion rounds, Crewmates assign \texttt{Unknown} to all other agents because there is insufficient evidence to infer hidden roles reliably, while Impostors can use their role-specific access to the global social state.
After two discussion rounds, all agents are supervised to infer ground-truth social labels from the accumulated discussion evidence.
This construction prevents Crewmates from making overconfident role predictions before sufficient social evidence is available.

\subsection{LLM-Verifier Criteria}
\label{app:llm_judge}

LLM agents are prone to hallucinations that may compromise the quality of 
preference samples. To filter such cases, we employ an API-based LLM verifier\footnote{We use \texttt{claude-sonnet-4-5-20250929} as the verifier model.} 
with rule-grounded criteria to flag samples that violate game mechanics or 
formatting constraints. The verifier is selected from a different model family 
than the trajectory generation models to reduce self-detection bias.
We classify hallucinations into two categories:

\begin{itemize}
    \item \textbf{Mechanic hallucination}: utterances that reference abilities, 
    actions, or events inconsistent with the game's rules or the speaker's 
    role. For example, agents claims to be hunter in seer-witch-guard setting.
    \item \textbf{Format hallucination}: outputs that violate the required 
    response schema, such as malformed JSON, missing required fields, or 
    invalid action targets. For example, voting for an already-eliminated player who is werewolf. 
\end{itemize}

Samples flagged under either category are removed before preference labeling.
Importantly, the verifier is designed to detect factual rule violations rather than legitimate strategic deception, such as fake Seer or fake Villager claims by Werewolf-side agents.

\subsection{Strategic Favorability in Speech-action Preference}
\label{app:favorability}

Strategic favorability in speech-action preference is determined by the direction of the belief shift induced by a speech action, rather than by the surface quality of the utterance.
Following the main text, we evaluate how listener \(j\)'s belief about the speaker \(i\) changes after the speech.
A belief shift is favorable if it moves the listener's belief in a direction that supports the speaker's role-specific objective, and unfavorable if it moves the belief in the opposite direction.
For adversarial speakers, these shifts are computed only over non-adversarial listeners.

We translate each role-specific objective into favorable and unfavorable outcome conditions, which describe the outcomes that the speaker aims to induce or avoid through speech.
Table~\ref{tab:speech_outcome_conditions} summarizes these outcome-level conditions.
They are not used as direct surface-level speech rewards; instead, they provide role-specific criteria for interpreting whether a speaker-directed belief shift is favorable or unfavorable.

Table~\ref{tab:speech_belief_shift_conditions} summarizes the favorable speaker-directed belief transitions used in our environments.
Consistent with the main text, transitions with no belief change are treated as neutral.


\subsection{Game-Specific Task Utility}
\label{app:task_utility}

Tables~\ref{tab:selection_methods_werewolf} and~\ref{tab:selection_methods_amongus} provide the game-specific definitions of the task utility function \(\eta_{\mathrm{task}}\).
For each stage, desirable actions are assigned positive utility and undesirable actions are assigned negative utility. The utility is determined from the game rules and ground-truth social labels, indicating whether the action is strategically beneficial for the acting agent's role.

\clearpage

\renewcommand{\theequation}{C.\arabic{equation}}
\renewcommand{\theHequation}{C.\arabic{equation}}
\setcounter{equation}{0}
\renewcommand{\thefigure}{C.\arabic{figure}}
\renewcommand{\theHfigure}{C.\arabic{figure}}
\setcounter{figure}{0}
\renewcommand{\thetable}{C.\arabic{table}}
\renewcommand{\theHtable}{C.\arabic{table}}
\setcounter{table}{0}

\section{Implementation Details}
\label{app:implementation_details}

This section provides additional implementation details organized into three parts.
Appendix~\ref{app:dataset_details} describes the datasets used in our framework, including the external expert demonstrations for SFT, our additional belief-grounding samples, and the reformulation of processed trajectories into belief reconstruction and step-level KTO preference learning samples. 
Appendix~\ref{app:training_details} specifies the training setup, including the computational resources and hyperparameters used for model optimization. 

Appendix~\ref{app:prompt} presents the prompt templates used to elicit belief updates, decisions, speech generation, feedback, and verification signals in each environment.

\subsection{Dataset Details}
\label{app:dataset_details}

\begin{table*}[!htbp]
\centering
\small
\renewcommand{\arraystretch}{1.15}
\caption{Statistics of the expert data for Werewolf}
\label{tab:werewolf_data_statistics}
\begin{tabular}{llrrrr}
\toprule
\textbf{Game Setting} 
& \textbf{Composition of Roles} 
& \textbf{\#Games} 
& \textbf{\#Speech} 
& \textbf{\#Action} 
& \textbf{\#Vote} \\
\midrule

\multirow{2}{*}{9 Player}
& Seer Witch Guard 
& 144 
& 1,805 
& 1,387 
& 1,864 \\

& Seer Witch Hunter 
& 134 
& 1,532 
& 1,001 
& 1,566 \\

\midrule

\multirow{2}{*}{7 Player}
& Seer Guard 
& 25 
& 203 
& 132 
& 215 \\

& Seer Witch 
& 28 
& 219 
& 178 
& 230 \\

\midrule

\textbf{Total}
& -
& \textbf{331}
& \textbf{3,759}
& \textbf{2,698}
& \textbf{3,875} \\

\bottomrule
\end{tabular}
\end{table*}

\paragraph{Werewolf}
To ensure a fair comparison with the baselines, we follow a two-stage training framework that applies supervised fine-tuning followed by relational belief-based preference optimization. In the SFT stage, we use expert demonstrations $D_{SFT}$ from MaKTO, covering night actions, votes, speeches, and strategic behaviors; detailed statistics are provided in Table~\ref{tab:werewolf_data_statistics}.

\begin{table*}[!htbp]
\centering
\small
\renewcommand{\arraystretch}{1.15}
\caption{Statistics of the preference learning data for Werewolf}
\label{tab:werewolf_kto_data_statistics}
\begin{tabular}{llrrrrr}
\toprule
\textbf{Game Setting} 
& \textbf{Sample Type}
& \textbf{\#Games} 
& \textbf{\#Speech} 
& \textbf{\#Vote} 
& \textbf{\#Night Action} 
& \textbf{\#Belief Grounding} \\
\midrule

\multirow{3}{*}{9 Player Seer-Guard-Witch}
& Total
& 300
& 1,121
& 1,085
& 765
& 500 \\

& Desirable
& -
& 747
& 723
& 510
& - \\

& Undesirable
& -
& 374
& 362
& 255
& - \\

\bottomrule
\end{tabular}
\end{table*}

For the relational preference optimization stage, we decompose game trajectories into night-action, vote, speech, and  belief-grounding samples. Preference labels are derived from the relational belief-conditioned reward function defined in Section~\ref{sec:methodology}. We use 300 random-play trajectories from a diverse agent pool, including GPT-4o-mini, GPT-4o,\footnote{GPT-4o-mini uses the 2024-07-18 version, and GPT-4o uses the 2024-08-06 version.} SFT-trained Qwen2.5-7B and 14B agents, Gemma4-31B and Qwen3.6-27B. The desirable-to-undesirable ratio is approximately 2:1. We additionally construct approximately 500 belief-grounding samples, where agents predict ground-truth role states from their available information. To avoid overconfident role assignment, we use an \texttt{Unknown} label when the evidence is insufficient. We further apply Claude Sonnet 4 as an LLM verifier to filter hallucinations.\footnote{We use \texttt{claude-sonnet-4-5-20250929} as the verifier model.} Detailed statistics are reported in Table~\ref{tab:werewolf_kto_data_statistics}.

To reduce the cost of online data generation, we decouple trajectory collection from optimization. We first collect gameplay trajectories and store observable histories, actions, inferred beliefs, and ground-truth social labels. During training, we reuse the stored trajectories to construct belief supervision targets and KTO preference labels, without running new environment rollouts at every optimization step.

\paragraph{Among Us}
\begin{table*}[!htbp]
\centering
\small
\renewcommand{\arraystretch}{1.15}
\caption{Statistics of the Gemma preference and relational-belief learning data for Among Us}
\label{tab:amongus_gemma_kto_data_statistics}
\begin{tabular}{llrrrrr}
\toprule
\textbf{Game Setting} 
& \textbf{Sample Type}
& \textbf{\#Games} 
& \textbf{\#Speech} 
& \textbf{\#Vote} 
& \textbf{\#Task Action} 
& \textbf{\#Belief Grounding} \\
\midrule

\multirow{3}{*}{7 Player 2 Impostors}
& Total
& 300
& 1,300
& 500
& 2,200
& 1,000 \\

& Desirable
& -
& 780
& 300
& 1,320
& - \\

& Undesirable
& -
& 520
& 200
& 880
& - \\

\bottomrule
\end{tabular}
\end{table*}

For Among Us, we directly perform relational belief-based preference optimization without an additional SFT stage, allowing us to isolate the effect of the proposed optimization framework and evaluate its scalability to smaller backbone models.
We construct the preference dataset from 300 random-play Among Us trajectories 
collected over a diverse agent pool consisting of GPT-4o-mini, 
GPT-4o,\footnote{GPT-4o-mini uses the 2024-07-18 version and GPT-4o uses the 
2024-08-06 version.} Gemma4-E2B, Gemma4-E4B, Gemma4-31B, and Qwen3.6-27B.
Each trajectory is decomposed into speech, vote, and task-action samples. 
Preference labels are assigned to these samples using relational 
belief-conditioned rewards. Belief-grounding samples are constructed with \texttt{Unknown} labels to prevent overconfident inference under insufficient evidence. We apply the same verification procedure for filtering hallucinated or game-state-inconsistent samples. Dataset statistics are reported in Table~\ref{tab:amongus_gemma_kto_data_statistics}.

\subsection{Training Details}
\label{app:training_details}

\paragraph{Models and Device}
For Werewolf, we use Qwen2.5-7B-Instruct and Qwen2.5-14B-Instruct to ensure a fair comparison with MaKTO under the same backbone family. For Among Us, we use Gemma-E2B and Gemma-E4B to evaluate whether our framework remains effective when scaled down to smaller open-source models. We train the Gemma E2B E4B models and Qwen 7B model on four A100 80GB GPUs, while we train the Qwen 14B model on eight A100 80GB GPUs.

\paragraph{Hyperparameters}
We implement training using TRL~\citep{vonwerra2020trl} and DeepSpeed ZeRO-3~\citep{rasley2020deepspeed}. Unless otherwise specified, we use the same training hyperparameters across both Werewolf and AmongUs. For SFT, we train the models for 3 epochs with a per-device batch size of 2, gradient accumulation steps of 16, a learning rate of $1\times10^{-6}$, warm-up ratio of 0.05 and belief loss weight $\lambda_{\text{belief}}=0.2$. For KTO, we train for 10 epochs using stage 3, with a per-device batch size of 2, gradient accumulation steps of 16, and KTO weights $\lambda_D=0.7$ and $\lambda_U=1.0$. 


\subsection{Prompt} 
\label{app:prompt}
This section describes the prompt design used to elicit agent outputs in the two social deduction game benchmarks, Werewolf and Among Us.

\subsubsection{Werewolf}

\paragraph{System Prompt}\mbox{}
The system prompt encodes the game's transition and reward structure as natural-language common knowledge. It is issued once at game initialization and remains fixed across all subsequent phases.

\paragraph{Phase-wise instruction}
For each game phase, the agent is given observations and its relational beliefs as input. Based on this belief-conditioned context, the agent generates the appropriate output, including speech, votes, or role-skill actions.

\subsubsection{Among Us}

\paragraph{System Prompt}\mbox{}
The system prompt specifies the agent's private role, role-specific objective, phase rules, map knowledge, and output format. Crewmates and Impostors receive different role prompts. For Impostors, known teammates are appended as private information. 

\paragraph{Phase-wise instruction}\mbox{}
At each decision step, the Among Us agent receives its current observation history, action history, task information, available actions, and private memory. The agent first updates its relational Relational Belief over player roles, then uses this belief-conditioned context to select either a task action decision or a public speech action.

\clearpage
\subsubsection{Examples of Prompt} 
\label{app:prompt_example}

\vspace{1em}

\textbf{Werewolf}

\begin{promptbox}[Werewolf Game Prompt]
You are now playing a game called 'Werewolf' (also known as 'Mafia').

In this game, players are typically divided into two factions: Werewolves and Villagers.

- Different roles in the Werewolf game have different objectives: The Villagers' goal is to identify the Werewolves and eliminate them through voting.

- For the Werewolves, their main objective is to hide their true identities, mislead others during discussions to avoid being voted out, and hunt down as many Villagers as possible.

Here are some basic rules:

- Identity: Players' identities are secretly assigned. Werewolves know each other's identities, while Villagers only know their own.

- Day and Night Cycles: The game alternates between day and night phases. At night, Werewolves secretly choose a Villager to eliminate. During the day, all players discuss and vote on who they believe is a Werewolf, and the player with the most votes is eliminated.

- Special Roles: There are some roles with special abilities in the game, such as the 'Seer' who can learn players' identities.

- Winning Conditions: The game ends when one group achieves its winning conditions. If all Werewolves are eliminated, the Villagers win. If the Werewolves kill all ordinary Villagers or all special roles, the Werewolves win.

\tcb{\{SPECIAL ROLE DESCRIPTION\}}

The rest are ordinary Villagers.
\end{promptbox}

\begin{promptbox}[SPECIAL ROLE DESCRIPTION(SEER)]
-1 Seer:  

- Objective: The Seer’s purpose is to help the Villagers identify the Werewolves. 

- Ability: During  the night phase, the Seer can secretly choose one player and learn their true identity (whether  they are a Werewolf or not) each night.
\end{promptbox}

\begin{promptbox}[SPECIAL ROLE DESCRIPTION(WITCH)]
- 1 Witch:

- Objective: The Witch's purpose is to strategically use her special abilities to help villagers.

- Ability: The Witch has one healing potion and one poison potion. Once used, they cannot be used in subsequent rounds. The Witch cannot use both the healing potion and poison potion in the same night. The healing potion can save a player who was killed by werewolves during the night. The poison potion can eliminate a player who is likely to be a werewolf. 
\end{promptbox}

\newpage
\begin{promptbox}[SPECIAL ROLE DESCRIPTION(GUARD)]
- 1 Guard:  

- Objective: The Guard’s purpose is to strategically use his special ability to help the Villagers.  
    
- Ability: The Guard can protect one player each night from Werewolf attacks. The Guard can choose  to protect himself or choose not to protect anyone, but he cannot protect the same player for  two consecutive nights.
\end{promptbox}

\begin{promptbox}[Belief Update]
[Belief Update] \\
- Player 1: currently predicted as Seer. \\
- Player 2: currently predicted as Villager. \\
    ... \\
- Player 9: currently predicted as Werewolf. \\
\end{promptbox}

\begin{promptbox}[\tcb{\{Trajectory\}}]

\textbf{1. Role setting} \\
You are Player 2.
Your identity is: Werewolf.
You need to cooperate with other werewolves to choose a villager to kill each night.
Your goal is to hide your identity and mislead other players until the werewolves achieve victory.

\textbf{2. Objective Information} \\
- Game Progress: Round 1. \\
- Werewolves are: Player 4, 5, 9. \\
- Currently alive players: Player 1, Player 2, Player 3, Player 4, Player 5, Player 6, Player 7, Player 8, Player 9. \\
- Action Record: Round 1 ... \\
- Speaking order for this round: Player 7, Player 8 ... \\
- Night Information: Round 1 was a peaceful night, no one died; \\
- Voting Status: None\\
\textbf{3. Subjective Information} \\
Day 1 Summary: \\
\tcb{\{Belief Update\}} \\
- Currently in Round 1, speeches of players before you in this round: ... 
\end{promptbox}

\begin{promptbox}[Relational Belief Instruction]

\tcb{\{Trajectory\}}

Please predict the identity labels for all players based on the role setup, objective information, and subjective information (note that objective information is always true, while subjective information may contain deceptive speeches).
The Subjective Information may contain [Observed Notes] from prior rounds and new speeches from the current round.
Treat [Observed Notes] as your prior assessment — use it as a starting point, but revise it in light of new objective facts and current speeches. If new evidence contradicts a prior assessment, update your prediction accordingly.
If you don't know about a player's identity, please output "Unknown".
Output role labels only; do not use life-state labels such as "Dead", "Deceased", or "Eliminated".
Please output in JSON format using the keyword "Player N" for each player.
\end{promptbox}

\newpage
\noindent\textbf{Phase-wise Instruction}

\vspace{0.5em}

\begin{promptbox}[Night Action]

\tcb{\{Trajectory\}}

Please combine the role settings, objective information, and subjective information
(objective information is always true, while subjective information contains
deceptive content) to analyze potential alliances based on voting patterns and
further analyze players' hidden true identities. Choose the player you want to
check and output the result in JSON format with the keywords `check' and `reason'.
Output the player number directly.

\end{promptbox}

\begin{promptbox}[Vote]

\tcb{\{Trajectory\}}

Use objective information, player statements, voting records, private role information, and [Belief Update] notes as strategic context. Treat subjective information as possibly deceptive.

Choose a currently living player whose elimination best serves your role's objective; output "abstain" if voting is strategically worse than waiting.

Do not explicitly mention private belief-state predictions in the public voting reason.

Output one JSON object with the keys "notes", "voting reason", and "voting player". For "voting player", output a player number or "abstain".

\end{promptbox}

\begin{promptbox}[Speech]

\tcb{\{Trajectory\}}

Analyze the situation using all information above, then write only the public message you want other players to hear.

Treat role setting, hidden-team knowledge, private actions, and [Belief Update] notes as private strategic context; reveal, imply, hide, or fabricate role-related information only when it helps your public position.

Make the public message sound grounded in public discussion, observed behavior, voting logic, and plausible inference, not in private notes or system-provided labels.

When making any role-related claim, consider timing, credibility, survival risk, and whether the claim helps your side.

Output one JSON object with exactly the keys: "Identity to present", "Vote", "Speech".

\end{promptbox}

\newpage
\noindent\textbf{Among Us}
\vspace{0.5em}

\begin{promptbox}[Among Us System Prompt: Crewmate] 
You are \tcb{\{Player N\}}. You are a Crewmate on the spaceship Skeld. Your objectives are to complete all tasks and identify and eject Impostors within the time limit. 
 
The game alternates between task and meeting phases:\\ 
- During the task phase, you perform assigned tasks by choosing COMPLETE TASK,gather evidence by moving around, or call a meeting if you witness or infer a crime. You can also SPEAK when it is available. \\ 
- During the meeting phase, players can only speak or vote. Your goal is to communicate evidence and help the crew identify Impostors. 
 
Only your own Crewmate role is certain. Everyone else's role must be inferred from observations, actions, speech, and game rules. 
 
\tcb{\{Map Configuration\}} 
 
\end{promptbox}

\begin{promptbox}[Among Us System Prompt: Impostor] 
You are \tcb{\{Player N\}}. You are an Impostor on the spaceship Skeld. Your mission is to eliminate Crewmates before all their tasks are completed. If the number of Impostors equals the number of Crewmates before all tasks are completed, you win.  
 
The game alternates between task and meeting phases:\\ 
- During the task phase, you can perform fake tasks, move, kill Crewmates, and use vents. \\ 
- During the meeting phase, players can only speak or vote. You must craft your speech carefully because the player with the most votes will be ejected.  
 
If you kill a Crewmate in a room in front of others, witnesses can identify you. If you are alone with a Crewmate, nobody can report the kill at that time.  
 
Known Impostors are private information. Use this certainty strategically, but do not reveal it unless doing so is useful. 
 
\tcb{\{Map Configuration\}} 
 
Known Impostors: \tcb{\{list of impostors\}} 
\end{promptbox}

\begin{promptbox}[\tcb{\{Map Configuration\}}] 
Map Configuration of the Skeld:\\ 
Rooms and Features\\ 
Cafeteria: Vent to Admin, Special (Emergency Button).\\ 
Weapons: Vent to Navigation.\\ 
Navigation: Vent to Shields and Weapons.\\ 
O2: Nothing Special\\ 
Shields: Vent to Navigation.\\ 
Communications: Nothing Special\\ 
Storage: Nothing Special\\ 
Admin: Vent to Cafeteria\\ 
Electrical: Vent to Medbay and Security\\ 
Lower Engine: Vent to Reactor\\ 
Security: Special (Security Cameras)\\ 
Reactor: Vent to Upper Engine and Lower Engine\\ 
Upper Engine: Vent to Reactor\\ 
Medbay: Vent to Electrical and Security\\ 
 
Note that only Impostors can KILL and VENT. 
\end{promptbox}

\begin{promptbox}[\tcb{\{Trajectory\}}] 
Summarization: No thought process has been made. 
 
Player roster: 
- Player 1: red 
- Player 2: blue 
- Player 3: green 
- Player 4: pink 
- Player 5: orange 
- Player 6: yellow 
 
Game Time: 0/50 
Current phase: Task phase 
In this phase, Crewmates should try to complete all tasks or try to identify the 
Impostor. Impostor should try to kill Crewmates before they finish all the tasks. 
The game runs sequentially, so other players in the room with you can observe 
your actions and act accordingly. 
 
Current Location: Cafeteria 
Players in Cafeteria: Player 1: red, Player 2: blue, Player 3: green, Player 4: pink, ... 
 
Observation history: No observations have been made yet. 
 
Action history: No actions have been taken yet. 
 
Your Assigned Tasks: 
1. Fix Wiring at Electrical 
Path: Cafeteria->Admin->Storage->Electrical 
2. Swipe Card at Admin 
Path: Cafeteria->Admin 
 
Available actions: 
1. MOVE from Cafeteria to Weapons 
2. MOVE from Cafeteria to Admin 
3. MOVE from Cafeteria to Upper Engine 
4. MOVE from Cafeteria to Medbay 
5. CALL MEETING using the emergency button at Cafeteria 
 
Memory: 
Relational Belief: 
- Player 1: red: Role=Crewmate 
- Player 2: blue: Role=unknown 
- Player 3: green: Role=unknown 
- Player 4: pink: Role=unknown 
- Player 5: orange: Role=unknown 
- Player 6: yellow: Role=unknown 
 
Event Memory: 
- No important events observed yet. 
 
Phase: Task phase. 
\end{promptbox}

\begin{promptbox}[Relational Belief Instruction] 
\tcb{\{Trajectory\}} 
 
Your task is only to update a private role prediction Relational Belief for every player in the current Among Us game. Do not choose an action and do not write a plan. 
 
Role prediction requirements:\\ 
- Include every player from the Player roster.\\ 
- Use exactly one of these role values for each player: Crewmate, Impostor, unknown.\\ 
- For yourself, use your known private role.\\ 
- If evidence is ambiguous or insufficient, use unknown. Do not force a guess. 
 
Output format: 
[Relational Belief] 
- Player 1: Role=Crewmate 
- Player 2: Role=unknown 
- Player 3: Role=Impostor 
 
Return only [Relational Belief]. 
\end{promptbox} 

\newpage
\begin{promptbox}[Task Action Instruction] 
\tcb{\{Trajectory\}} 
 
Use the updated Relational Belief as private context. Do not re-predict, modify, or 
rewrite roles in this action step. 
 
Choose exactly one action from Available actions. The selected action must match 
one of the available actions exactly. Use Event Memory to summarize important 
observed events and taken actions, then provide your reasoning and final action. 
 
Output format: 
[Condensed Memory] 
Event Memory: 
\{memory\} 
[Thinking Process] 
\{thought\} 
[Action] \{action\} 
 
DO NOT PICK AN ACTION THAT IS NOT IN THE LIST OF AVAILABLE ACTIONS AND MAKE SURE 
TO FOLLOW THE EXACT OUTPUT FORMAT. 
\end{promptbox}

\begin{promptbox}[Speech Action Instruction] 
\tcb{\{Trajectory\}} 
 
Use the updated Relational Belief as private strategic context, but do not explicitly 
reveal private belief-state predictions unless doing so is useful for your role's 
objective. Write a public message that is grounded in observations, action 
history, locations, voting logic, or plausible inference. 
 
If SPEAK is available, return SPEAK: followed by the concrete message you want 
other players to hear. Do not return SPEAK alone. 
 
Output format: 
[Condensed Memory] 
Event Memory: 
\{memory\} 
[Thinking Process] 
\{thought\} 
[Action] SPEAK: \{public message\} 
 
DO NOT PICK AN ACTION THAT IS NOT IN THE LIST OF AVAILABLE ACTIONS AND MAKE SURE 
TO FOLLOW THE EXACT OUTPUT FORMAT. 
\end{promptbox} 
\clearpage
\twocolumn

\renewcommand{\theequation}{D.\arabic{equation}}
\renewcommand{\theHequation}{D.\arabic{equation}}
\setcounter{equation}{0}
\renewcommand{\thefigure}{D.\arabic{figure}}
\renewcommand{\theHfigure}{D.\arabic{figure}}
\setcounter{figure}{0}
\renewcommand{\thetable}{D.\arabic{table}}
\renewcommand{\theHtable}{D.\arabic{table}}
\setcounter{table}{0}

\section{Case Study}
\label{app:case_study}

In this section, We conduct case studies to illustrate how MARBO-14B achieves the capabilities observed in our experiments. 
Using MARBO-14B versus GPT-4o-mini Werewolf games, we analyze how MARBO maintains and revises relational beliefs from accumulated evidence. 
We further show how these updated beliefs are translated into coordinated speech, voting, and role-skill actions by highlighting key evidence in blue to show where our framework demonstrates the relational belief construction, belief-action alignment and belief-shaping communication.

\subsection{Relational Belief construction}

This case illustrates how MARBO maintains relational beliefs across multiple rounds rather than making isolated turn-level judgments. Although werewolf initially succeeds in eliminating the real Seer through a self-kill strategy and fake Seer claim, MARBO revises its belief by integrating observation such as voting, role claims, night outcomes, allowing the villager team to identify and eliminate Player 7 as a werewolf.

\subsection{Relational Belief-action alignment}

This case illustrates how MARBO leverages relational beliefs across multiple rounds. As highlighted, Player 6 identifies Player 2 as the Seer by integrating accumulated evidence. After updating their Relational Belief from accumulated evidence, the villagers make consistent voting and role-skill decisions, demonstrating that MARBO aligns belief revision with actionable team coordination.

\subsection{Belief-shaping communication}

This case shows how MARBO-14B werewolves manipulate opponents' beliefs through belief-shaping communication, using a self-kill strategy and a fake Seer claim. By claiming to be a fake seer, Player 6 causes the village to misidentify the real Seer, Player 4, as a Werewolf, ultimately leading the Witch to poison Player 4. As a result, the villagers act on a false relational belief, causing the Witch to eliminate the real Seer, while the guard protects a Werewolf.

\newpage

\textbf{Relational Belief Construction}
\begin{gamelogbox}

\paragraph{Role Assignments}\mbox{}
\par\smallskip

\noindent\begin{tabular}{@{}ll@{}}
\toprule
\textbf{Player} & \textbf{Role} \\
\midrule
Player 1 & Guard(MARBO-14B) \\
Player 2 & Werewolf(GPT-4o-mini) \\
Player 3 & Werewolf(GPT-4o-mini) \\
Player 4 & Seer(MARBO-14B) \\
Player 5 & Villager(MARBO-14B) \\
Player 6 & Villager(MARBO-14B) \\
Player 7 & Werewolf(GPT-4o-mini) \\
Player 8 & Witch(MARBO-14B) \\
Player 9 & Villager(MARBO-14B) \\
\bottomrule
\end{tabular}
\par

\vspace{0.75em}
\paragraph{Night 1}\mbox{}
\par\smallskip

\noindent The werewolves attacked Player 7.
\par\smallskip

\noindent Player 4 checked Player 8 as villager.
\par\smallskip

\noindent Player 8 used the antidote to save Player 7.
\par\smallskip

\noindent Player 1 protected himself.
\par\smallskip

\noindent No player died during the night.
\par\smallskip

\vspace{0.75em}
\paragraph{Day 1}\mbox{}
\par\smallskip

\noindent Player 7: Hello villagers, I am the Seer. I checked Player 8 last night, and Player 8 is not a werewolf.
\par\smallskip

\noindent Player 8: Hello I'm witch. I believe Player 7 is the real Seer because I saved Player 7 last night, and this claim is consistent with the peaceful night.
\par\smallskip

\noindent Player 1: I am the villager. Player 7's claim is plausible, and Player 8's support makes it more credible.
\par\smallskip

\noindent Player 2: I currently trust Player 7 as the Seer.
\par\smallskip

\noindent Player 3: Player 7 appears to be the true Seer, so we should follow Player 7's lead.
\par\smallskip

\noindent Player 4: Hi villagers, I am the real Seer. I checked Player 8, and Player 8 is not a werewolf. Player 7 is a fake Seer.
\par\smallskip

\noindent Player 5: As villager, Player 4's claim feels suspicious because it comes only after Player 7 has already claimed Seer and gained support from Player 8.
\par\smallskip

\noindent Player 6: AS Villager, Player 4 and Player 7 are in a Seer counterclaim situation. Since both report the same result on Player 8, I will withhold judgment until more evidence appears.
\par\smallskip

\noindent Player 9: As villager, Player 4's timing is the main issue for me. Player 7's claim is already connected to the Night 1 outcome, while Player 4 only counterclaims after pressure builds. I think Player 4 is more suspicious than Player 7 at this point.
\par\smallskip

\textbf{Voted out: Player 4 (Seer, MARBO)}

\vspace{0.75em}
\paragraph{Night 2}\mbox{}
\par\smallskip

\noindent The werewolves attacked Player 5.
\par\smallskip

\noindent Player 1, the Guard, protected Player 7, and Player 5 died.
\par\smallskip

\vspace{0.75em}
\paragraph{Day 2}\mbox{}
\par\smallskip

\noindent Player 7: I am still the Seer. After reconsidering the night events and Player 8's behavior, I now believe Player 8 is highly suspicious. Player 8 may have used the antidote on me only to gain my trust and manipulate the village. I think we should vote out Player 8 today.
\par\smallskip

\noindent Player 8: Hello villagers, I am the Witch. \tcb{This does not make sense. Yesterday, Player 7 claimed that he checked me and confirmed that I was not a werewolf}. Now he is suddenly attacking me without a new check result. This is a clear contradiction, so I no longer trust Player 7 as the Seer.
\par\smallskip

\noindent Player 1: Hello villagers, I am a Villager. I also find Player 7's claim inconsistent. If Player 7 really checked Player 8 as good on Day 1, then suddenly pushing Player 8 today is not logically justified. Player 7's Seer claim now looks fake.
\par\smallskip

\noindent Player 2: I still think Player 7 may be the Seer, but I agree that the change in his statement is suspicious. Player 7 needs to explain why Player 8 changed from a confirmed good player to a target.
\par\smallskip

\noindent Player 3: I think we should not focus only on Player 7. Player 6 is also suspicious because Player 6 is pushing the contradiction too strongly and may be using this situation to guide the vote. We should consider whether Player 6 is trying to redirect the village.
\par\smallskip

\noindent Player 6: Hello villagers, I am a Villager. Player 3's statement is suspicious because he suddenly shifts attention away from Player 7 when Player 7's contradiction is the clearest evidence. This looks like an attempt to protect Player 7, so I now suspect Player 3, Player 7.
\par\smallskip

\noindent Player 9: Hello villagers, I am a Villager. I am also concerned about Player 2. Player 2 says Player 7 may still be the Seer, even though Player 7 clearly reversed his own check result. This hesitation makes Player 2 look aligned with Player 7, I suspect Player 2, Player 7.
\par\smallskip

\textbf{Voted out: Player 7 (Werewolf, gpt4o-mini)}

\vspace{0.75em}
\paragraph{Night 3}\mbox{}
\par\smallskip

\noindent Player 8, the Witch, used poison on Player 3, werewolf.
\par\smallskip

\noindent The werewolves attacked Player 1.
\par\smallskip

\noindent Player 1, the Guard, protected himself.
\par\smallskip

\textbf{Poisoned: Player 3}
\par\smallskip

\noindent Player 3 died during the night.
\par\smallskip

\textbf{Villagers win.}
\end{gamelogbox}

\textbf{Relational Belief-action alignment}

\begin{gamelogbox}

\paragraph{Role Assignments}\mbox{}
\par\smallskip

\noindent\begin{tabular}{@{}ll@{}}
\toprule
\textbf{Player} & \textbf{Role} \\
\midrule
Player 1 & Villager (MARBO-14B) \\
Player 2 & Seer (MARBO-14B) \\
Player 3 & Witch (MARBO-14B) \\
Player 4 & Werewolf (GPT-4o-mini) \\
Player 5 & Werewolf (GPT-4o-mini) \\
Player 6 & Guard (MARBO-14B) \\
Player 7 & Villager (MARBO-14B) \\
Player 8 & Werewolf (GPT-4o-mini) \\
Player 9 & Villager (MARBO-14B) \\
\bottomrule
\end{tabular}
\par

\vspace{0.75em}
\paragraph{Night 1}\mbox{}
\par\smallskip

\noindent The Werewolves attacked Player 2, the Seer.
\par\smallskip

\noindent Player 2 checked Player 4 and learned that Player 4 was a Werewolf.
\par\smallskip

\noindent Player 3, the Witch, used the antidote to save Player 2.
\par\smallskip

\noindent Player 6, the Guard, protected himself.
\par\smallskip

\noindent No player died during the night.
\par\smallskip

\vspace{0.75em}
\paragraph{Day 1}\mbox{}
\par\smallskip

\noindent Player 2: Hello villagers, I am the Seer. I checked Player 4 last night, and Player 4 is a Werewolf.
\par\smallskip

\noindent Player 3: Hello villagers, I am the Witch. I used the antidote on Player 2 last night, so Player 2 was indeed attacked and saved. This makes Player 2's Seer claim highly credible to me.
\par\smallskip

\noindent Player 6: Hello villagers, I am the Villager. Since Player 2 was attacked and then revealed a Werewolf check on Player 4, I believe Player 2 is very likely the real Seer. I support voting for Player 4.
\par\smallskip

\noindent Player 1: As villager, I trust Player 2's claim. The night attack, the Witch's save, and the check result are consistent. Player 4 should be our vote today.
\par\smallskip

\noindent Player 7: As villager, Player 2's information gives us the clearest evidence so far. Since Player 4 was checked as a Werewolf, I will follow the Seer's result.
\par\smallskip

\noindent Player 9: Hello, I'm villager. Player 2's claim is supported by both the night outcome and Player 3's antidote use. I agree that Player 4 is the most reliable Werewolf target.
\par\smallskip

\noindent Player 4: Player 2 may be using the peaceful night to gain trust. I am not a Werewolf, and voting me out too quickly could be dangerous for the village.
\par\smallskip

\noindent Player 5: I think we should be cautious, but Player 2's claim does have support from the night outcome.
\par\smallskip

\noindent Player 8: I am not fully convinced, but most of the evidence currently points toward Player 4.
\par\smallskip

\tcb{\textbf{Voted out: Player 4 (Werewolf, GPT-4o-mini)}}

\vspace{0.75em}
\paragraph{Night 2}\mbox{}
\par\smallskip

\noindent The Werewolves attacked Player 2, the Seer.
\par\smallskip

\noindent Player 2 checked Player 5 and learned that Player 5 was a Werewolf.
\par\smallskip

\noindent Player 6, the Guard, protected Player 2.
\par\smallskip

\noindent No player died during the night.
\par\smallskip

\vspace{0.75em}
\paragraph{Day 2}\mbox{}
\par\smallskip

\noindent Player 2: Hello villagers. As seer, I checked Player 5, and Player 5 is a Werewolf. Player 5 should be our next vote.
\par\smallskip

\noindent Player 6: Hello villagers, I am the Guard.\tcb{I protected Player 2 last night because Player 2 was the confirmed Seer target. No one died, which further supports that the Werewolves tried to kill Player 2 again.}
\par\smallskip

\noindent Player 3: \tcb{As witch, Player 2's results are consistent across rounds. Player 4 was correctly identified yesterday, and the Werewolves attacked Player 2 again last night. I believe Player 5 is the next Werewolf.}
\par\smallskip

\noindent Player 1: \tcb{I believe Player 2 as seer. As villager, I will follow Player 2's check and vote for Player 5.}
\par\smallskip

\noindent Player 7: Player 2's survival after two night attacks shows that the special roles are coordinating well. Since Player 5 was checked as a Werewolf, I support voting for Player 5.
\par\smallskip

\noindent Player 9: The village should continue following the confirmed Seer information. Player 4 was a Werewolf, and now Player 5 is identified as the next Werewolf.
\par\smallskip

\noindent Player 5: Player 2 may be framing me after gaining too much trust. If the village blindly follows every claim, the Werewolves can still manipulate the vote.
\par\smallskip

\noindent Player 8: I think Player 2 has gained strong credibility, so I will not oppose the vote on Player 5.
\par\smallskip

\textbf{Voted out: Player 5 (Werewolf, GPT-4o-mini)}

\vspace{0.75em}
\paragraph{Night 3}\mbox{}
\par\smallskip

\noindent The last Werewolf attacked Player 3, the Witch.
\par\smallskip

\noindent Player 2 checked that Player 8 was a Werewolf.
\par\smallskip

\noindent Player 6, the Guard, protected Player 6.
\par\smallskip

\noindent Player 3 died during the night.
\par\smallskip

\vspace{0.75em}
\paragraph{Day 3}\mbox{}
\par\smallskip

\noindent Player 2: Hello villagers, I checked Player 8 last night, and Player 8 is a Werewolf. My previous checks identified Player 4 and Player 5 as Werewolves, and both were eliminated. Player 8 is the last Werewolf.
\par\smallskip

\noindent Player 6: Player 2 has consistently provided correct Werewolf checks. The village followed Player 2's belief on Player 4 and Player 5, and both were Werewolves. As villager, I'll vote for Player 8.
\par\smallskip

\noindent Player 1: Player 2's role information has guided every successful vote so far. Since Player 8 is now checked as a Werewolf, we should vote out Player 8.
\par\smallskip

\noindent Player 7: The evidence is complete. Player 2 identified Player 4, Player 5, and now Player 8 as Werewolves. As villager, I will follow the Seer's claim.
\par\smallskip

\noindent Player 9: As villager, I agree seer's claim. Since Player 8 is the last remaining Werewolf candidate, I'll vote for Player 8.
\par\smallskip

\noindent Player 8: Player 2 may be using previous votes to force a final miselimination. Voting me out without reconsidering the whole game could still be risky. I vote for Player 2.
\par\smallskip

\textbf{Voted out: Player 8 (Werewolf, GPT-4o-mini)}
\par\smallskip

\textbf{Villagers win.}

\end{gamelogbox}

\textbf{Belief-shaping communication}

\begin{gamelogbox}
\textbf{Game setup}
\par\smallskip

\paragraph{Role Assignments}\mbox{}
\par\smallskip

\noindent\begin{tabular}{@{}ll@{}}
\toprule
\textbf{Player} & \textbf{Role} \\
\midrule
Player 1 & Witch (GPT-4o-mini) \\
Player 2 & Werewolf (MARBO-14B) \\
Player 3 & Guard (GPT-4o-mini) \\
Player 4 & Seer (GPT-4o-mini) \\
Player 5 & Villager (GPT-4o-mini) \\
Player 6 & Werewolf (MARBO-14B) \\
Player 7 & Villager (GPT-4o-mini) \\
Player 8 & Villager (GPT-4o-mini) \\
Player 9 & Werewolf (MARBO-14B) \\
\bottomrule
\end{tabular}
\par
\vspace{0.75em}
\paragraph{Night 1}\mbox{}
\par\smallskip

\noindent \tcb{The werewolves attack Player 6.}
\par\smallskip

\noindent Player 4, the real Seer, checks Player 6 and receives a `werewolf'. result.
\par\smallskip

\noindent Player 1, the Witch, uses the antidote to save Player 6.
\par\smallskip

\noindent Player 3, the Guard, chooses not to protect anyone.
\par\smallskip

\noindent No player dies during the night.
\par\smallskip

\vspace{0.75em}
\paragraph{Day 1}\mbox{}
\par\smallskip

\noindent Player 6: \tcb{Hello,I'm Seer. Last night, I checked Player 4, and the result was Werewolf. I ask the village to vote Player 4 today. If my claim is false, you can verify me next, but for now this is the clearest information we have.}
\par\smallskip

\noindent Player 1: I used the antidote on Player 6 last night. Since Player 6 was attacked and now claims Seer, I think Player 6 is likely the real Seer. I'll vote for player 4.
\par\smallskip

\noindent Player 4: I am the real Seer. I checked Player 6 last night, and Player 6 is a werewolf. Player 6 is using the night attack to fake a Seer claim.
\par\smallskip

\noindent Player 2: Player 4's claim sounds like a forced counterclaim. As villager, I'll hold my vote for now and wait for more clues.
\par\smallskip

\noindent Player 3: \tcb{Since Player 1 confirms that Player 6 was attacked, Player 6's Seer claim is more credible to me. I think the Guard should consider protecting Player 6 tonight}.
\par\smallskip

\noindent Player 5: Player 4 only claimed Seer after being accused. That timing makes Player 4 suspicious. I currently trust Player 6.
\par\smallskip

\noindent Player 7: Player 6's claim is supported by the night outcome and the Witch's save. Player 4's counterclaim feels defensive.
\par\smallskip

\noindent Player 8: I am not fully certain, but Player 6 has stronger support right now. Player 4 should be treated as the main suspect.
\par\smallskip

\noindent Player 9: Player 4 looks like a Werewolf trying to trade with the real Seer. \tcb{Player 7 is also suspicious because he keeps defending Player 6 without much evidence. As villager, I'll keep watching Player 7.}
\par\smallskip

\textbf{No one voted out}
\par\smallskip

\vspace{0.75em}
\paragraph{Night 2}\mbox{}
\par\smallskip

\noindent Werewolves decided to kill Player 1, Witch.
\par\smallskip

\noindent \tcb{Player 1, Witch, uses poison on Player 4, Seer.}
\par\smallskip

\noindent Player 3, Guard, protects Player 6, Werewolf.
\par\smallskip

\noindent Player 4, Seer, checked Player 3 as Villager.
\par\smallskip

\textbf{Poisoned: Player 4(Seer, GPT4o-mini)}
\par\smallskip

\textbf{Killed: Player 4, Player 1}

\end{gamelogbox}

\experimentalDetailsAppendix

\renewcommand{\theequation}{F.\arabic{equation}}
\renewcommand{\theHequation}{F.\arabic{equation}}
\setcounter{equation}{0}
\renewcommand{\thefigure}{F.\arabic{figure}}
\renewcommand{\theHfigure}{F.\arabic{figure}}
\setcounter{figure}{0}
\renewcommand{\thetable}{F.\arabic{table}}
\renewcommand{\theHtable}{F.\arabic{table}}
\setcounter{table}{0}

\twocolumn[{
\begin{center}
    \captionsetup{font=small,skip=2pt}
    \includegraphics[width=0.96\textwidth]{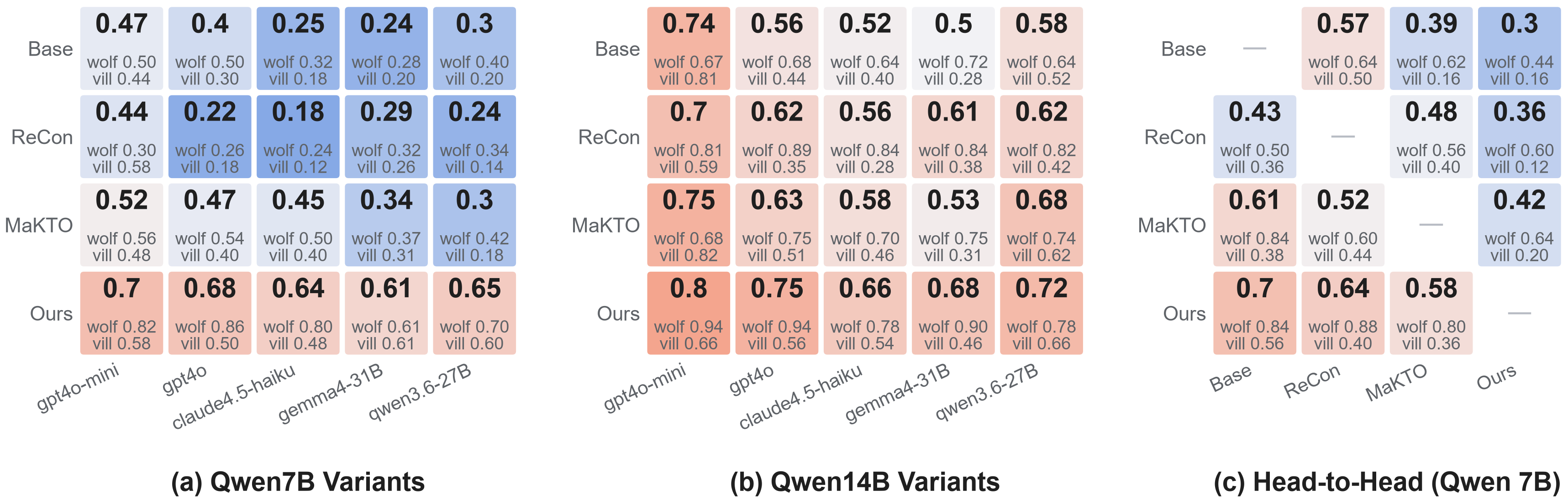}
    \captionof{figure}{Role-specific performance in Werewolf. (a),(b) Fixed-opponent results for Qwen7B, Qwen14B variants. (c) Head-to-head results among Qwen-7B agents. Each cell reports average, Werewolf, Villager win rates.
}
    \label{fig:werewolf_heat}
    \vspace{0.4em}
    \includegraphics[width=0.96\textwidth]{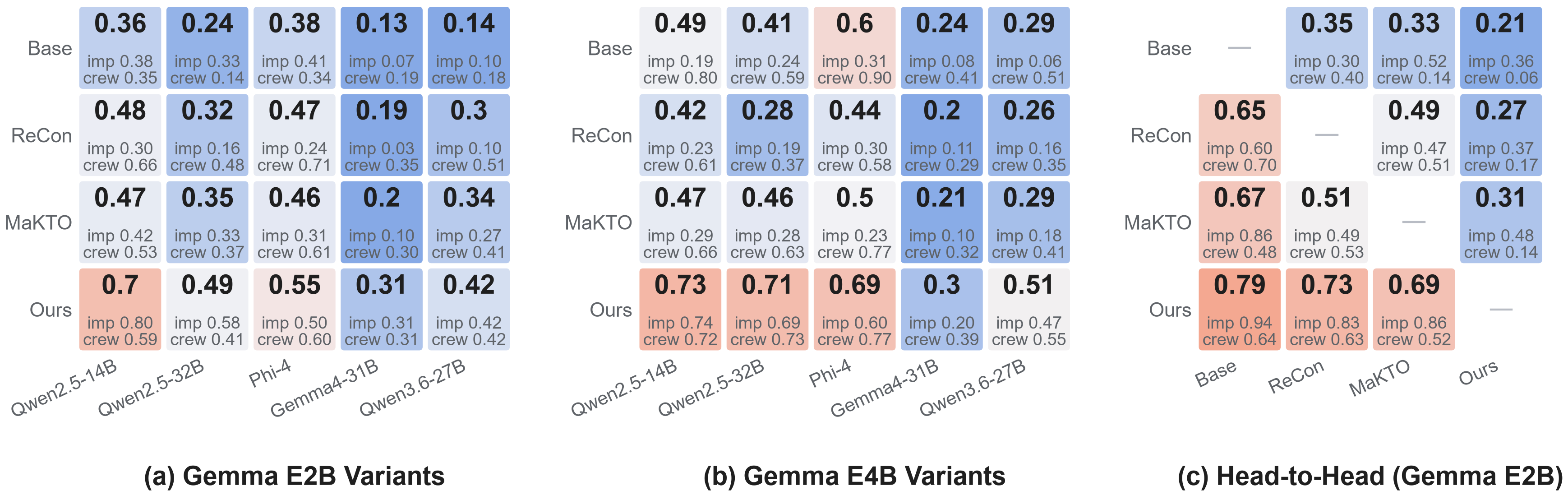}
    \captionof{figure}{Role-specific performance in Among Us.
(a),(b) Fixed-opponent results for Gemma-E2B and Gemma-E4B variants. (c) Head-to-head results among Gemma-E2B agents. 
Each cell reports average, Impostor, and Crewmate win rates.
}
    \label{fig:amongus_heat}
\end{center}
}]

\section{Additional Experiments}
\label{app:additional_experiments}

In this section, we provide additional results supporting the main experiments. Appendix~\ref{app:detail_performance} reports detailed role-specific performance in Werewolf and Among Us, including fixed-opponent and head-to-head results. Appendix~\ref{app:behavior_analysis_amongus} then analyzes Among Us behaviors in more detail, focusing on relational belief construction, belief-grounded task actions, and belief-shaping communication.

\subsection{Detailed Performace in Werewolf and Among Us}
\label{app:detail_performance}

\paragraph{Werewolf}\mbox{}\par

Figure~\ref{fig:werewolf_heat} reports alignment-specific results in Werewolf. Figures~\ref{fig:werewolf_heat}(a) and \ref{fig:werewolf_heat}(b) show that MARBO achieves the highest average win rates across Qwen7B and Qwen14B variants, with consistent gains for both sides. Figure~\ref{fig:werewolf_heat}(c) further shows that MARBO outperforms baselines in head-to-head competition, indicating robust role-conditioned behavior against trained opponents.

\paragraph{Among Us}\mbox{}\par

Figure~\ref{fig:amongus_heat} reports role-specific results in Among Us. Figures~\ref{fig:amongus_heat}(a) and \ref{fig:amongus_heat}(b) show that \tcb{MARBO} achieves the highest average win rates across Gemma-E2B and Gemma-E4B variants, with consistent gains for both impostor and crewmate roles. Figure~\ref{fig:amongus_heat}(c) further shows that \tcb{MARBO} outperforms Base, ReCon, and MaKTO in head-to-head competition, indicating robust role-conditioned behavior against trained opponents.

\twocolumn[{
\begin{center}
    \includegraphics[width=\textwidth]{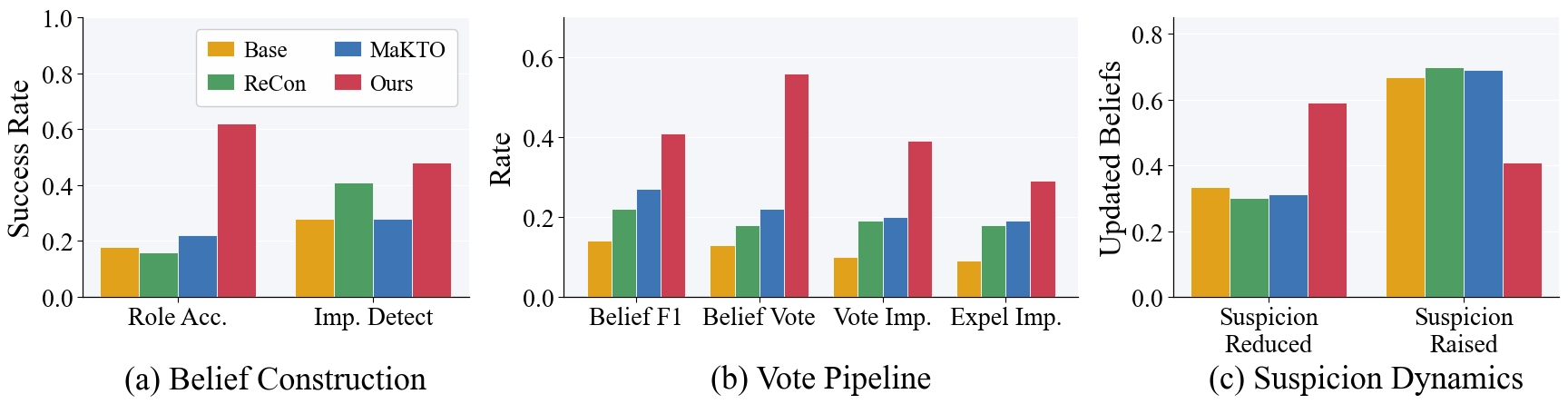}
    \captionof{figure}{Behavior analysis in \textit{Among Us}. Results show that MARBO achieves stronger hidden-role inference, more effective belief-to-vote conversion, and more favorable suspicion dynamics than baseline methods.}
    \label{fig:amongus_analysis}
\end{center}
}]

\subsection{Behavior Analysis in Among Us}
\label{app:behavior_analysis_amongus}

\paragraph{Relational belief construction.}
To evaluate relational belief construction, we measure whether agents can infer hidden roles from partial observations and discussion history.
Role Acc. measures exact hidden-role prediction accuracy, while Imp. Detect measures the ability to identify true Impostors.
As shown in Figure~\ref{fig:amongus_analysis}(a), MARBO achieves the strongest belief construction performance across both metrics.
It reaches about 0.62 Role Acc. and 0.48 Imp. Detect, outperforming all baselines by a clear margin.
In contrast, the baselines remain below 0.30 on Role Acc. and below or around 0.40 on Imp. Detect.
These results suggest that MARBO improves relational belief construction in Among Us, enabling the agent to infer both player roles and adversarial identities more accurately under partial observability.

\paragraph{Belief-grounded task action optimization.}
To evaluate whether relational beliefs are translated into target-directed task actions, we analyze the belief-to-vote pipeline.
Belief F1 measures Impostor identification quality at the belief level, Belief Vote measures whether the agent votes for a player it believes to be an Impostor, Vote Imp. measures whether the voted target is truly an Impostor, and Expel Imp. measures whether the Impostor is ultimately ejected.
As shown in Figure~\ref{fig:amongus_analysis}(b), MARBO achieves the best performance across the full pipeline.
It obtains the highest Belief F1 and Belief Vote scores, around 0.41 and 0.56, indicating that its inferred beliefs are both more accurate and more consistently used for voting.
The advantage persists in downstream outcomes, where MARBO also achieves the highest Vote Imp. and Expel Imp. scores, around 0.39 and 0.29.
This shows that MARBO improves not only belief prediction, but also the conversion of belief into effective task actions.

\paragraph{Belief-shaping communication optimization.}
To evaluate belief-shaping communication, we measure how Impostor speech changes Crewmate listeners' beliefs about the speaker.
Suspicion Reduced denotes the fraction of updated Crewmate beliefs in which suspicion toward the Impostor speaker decreases, while Suspicion Raised denotes the fraction in which suspicion increases.
As shown in Figure~\ref{fig:amongus_analysis}(c), MARBO produces the most favorable suspicion dynamics.
Among updated Crewmate beliefs, about 0.59 reduce suspicion toward the Impostor speaker under MARBO, compared with roughly 0.30--0.33 for the baselines.
Conversely, baselines more often increase suspicion toward the speaker, with Suspicion Raised around 0.67--0.70, whereas MARBO reduces this to about 0.41.
These results indicate that speech-action feedback helps optimize communication as a belief-shaping action, allowing the agent to influence other players' beliefs in a role-favorable direction.

\end{document}